\documentclass[conference]{IEEEtran}

\usepackage[T1]{fontenc}
\usepackage{cite}
\usepackage{amsmath,amssymb,amsfonts}
\usepackage{algorithmic}
\usepackage{graphicx}
\usepackage{textcomp}
\usepackage{xcolor}
\usepackage{algorithm2e}
\usepackage{booktabs}
\usepackage{caption}
\usepackage[disable]{todonotes}
\usepackage{makecell}
\usepackage{balance}
\usepackage{hyperref}

\hypersetup{colorlinks=true,urlcolor=blue}
\IEEEoverridecommandlockouts

\begin{document}

\title{Combating noisy labels in object detection datasets
%\thanks{This work was supported by the National Centre for Research and Development under the project LIDER/51/0221/L-
%11/19/NCBR/2020.}
}

\author{\IEEEauthorblockN{1\textsuperscript{st} Krystian Chachuła}
\IEEEauthorblockA{\textit{Silesian University of Technology}\\
Gliwice, Poland \\
krystian@krystianch.com}
\and
\IEEEauthorblockN{2\textsuperscript{nd} Jakub Łyskawa}
\IEEEauthorblockA{\textit{Warsaw University of Technology}\\
Warsaw, Poland \\
jakub.lyskawa.dokt@pw.edu.pl}
\and
\IEEEauthorblockN{3\textsuperscript{rd} Bartłomiej Olber}
\IEEEauthorblockA{\textit{Warsaw University of Technology}\\
Warsaw, Poland \\
bartlomiej.olber.stud@pw.edu.pl}
\and
\IEEEauthorblockN{4\textsuperscript{th} Piotr Frątczak}
\IEEEauthorblockA{\textit{Warsaw University of Technology}\\
Warsaw, Poland \\
piotr.fratczak.stud@pw.edu.pl}
\and
\IEEEauthorblockN{5\textsuperscript{th} Adam Popowicz}
\IEEEauthorblockA{\textit{Silesian University of Technology}\\
Gliwice, Poland \\
adam.popowicz@polsl.pl}
\and
\IEEEauthorblockN{6\textsuperscript{th} Krystian Radlak}
\IEEEauthorblockA{\textit{Warsaw University of Technology}\\
Warsaw, Poland \\
krystian.radlak@pw.edu.pl}
}

\thispagestyle{empty}

\twocolumn[{%
\renewcommand\twocolumn[1][]{#1}%
\maketitle
\begin{center}
    \centering
    \captionsetup{type=figure}
    \includegraphics[width=\textwidth]{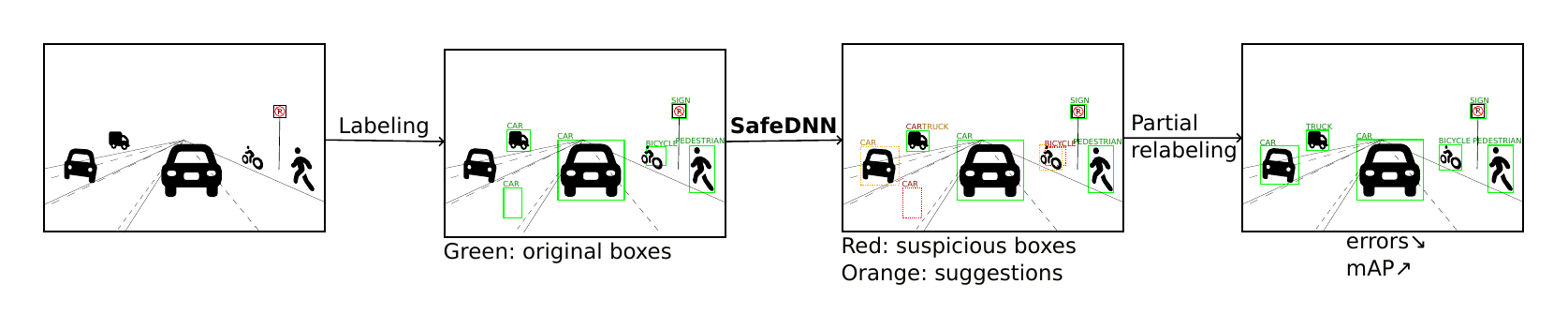}
    \vspace*{-25pt}
    \captionof{figure}{A proposed use-case of the CLOD algorithm.}
    \vspace{25pt}
    \label{fig:meme}
\end{center}%
}]

\newcommand\imw{0.7\linewidth}

\begin{abstract}
The quality of training datasets for deep neural networks is a key factor
contributing to the accuracy of resulting models.
This effect is amplified in difficult tasks such as object detection.
Dealing with errors in datasets is often limited to accepting
that some fraction of examples are incorrect, estimating their confidence,
and either assigning appropriate weights or ignoring uncertain ones during training.
In this work, we propose a different approach.
We introduce the Confident Learning for Object Detection (CLOD) algorithm
for assessing the quality of each label in object detection datasets,
identifying missing, spurious, mislabeled, and mislocated bounding boxes
and suggesting corrections.
By focusing on finding incorrect examples in the training datasets, we
can eliminate them at the root.
Suspicious bounding boxes can be reviewed to improve the quality
of the dataset, leading to better models without further complicating their
already complex architectures.
The proposed method is able to point out nearly 80\% of artificially disturbed
bounding boxes with a false positive rate below 0.1.
Cleaning the datasets by applying the most confident automatic suggestions improved mAP scores by 16\%
to 46\%, depending on the dataset, without any modifications to the network architectures.
This approach shows promising potential in rectifying state-of-the-art object detection datasets.
\end{abstract}


% ------------------------------------------------------------------------------ <- (80 cols)

\section{Introduction}

Machine learning methods have been widely used for image analysis in recent
years.
Deep learning (DL), specifically deep neural networks (DNNs), are the typical choice for solving these tasks.
Generally speaking, DNN models outperform competing deterministic algorithms.
However, the quality of the training datasets has a significant impact on
the results of DL algorithms \cite{Popowicz2022CombatingLN}.
The eventual correctness of the algorithm depends on the size and quality of such datasets,
which can contain hundreds of thousands of annotated images.
Due to the importance of such datasets in different sectors, various
organizations create them through the collection and tagging of
data.

The bulk of training datasets that are freely accessible may only be utilized
for non-commercial purposes.
Therefore, enterprises must often prepare them on their own.
The created datasets are labeled either automatically or by
crowdsourcing \cite{ChangAmershi2017}.
Frequently, enterprises hire a third party to verify the label's quality.
This third party is often made up of inexperienced workers, so
the accuracy of tagging is lacking.
Willers et al. \cite{WillersSudholt2020} demonstrated that the labeling quality is
one of the primary reasons for the reliability decline of DL algorithms.
We also assessed the influence of label quality on model quality (see Sec.~\ref{sec:noisydata}).

There are many causes of labeling errors, also referred to as \textit{label noise}.
For example, insufficient information in the images to properly identify objects or
different opinions of individual annotators \cite{Frnay2014ACI}. 
In such cases, detecting label noise may be
a good first step to improving the instructions for annotating images, thus contributing to the higher quality of resulting datasets.

The usage of DL algorithms in safety-critical applications (especially for object detection problems in autonomous driving vehicles) requires proving that the training
dataset is free of defects and legal constraints. The quality of the training dataset
needs to be assured, and therefore, formal methods for proving the
trustworthiness of training data need to be included in the development life cycle of object detectors.
Moreover, the numerous safety concerns for AI systems point to the necessity of
assessing the labeling accuracy of training datasets, but the existing safety standards very weakly address the problem of data labeling quality.
ISO/DIS 21448 \cite{ISO-21448} requires identifying the possible source of limitations for each stage of
development of machine learning models and defining corresponding countermeasures. However, the standard recommends only performing a
review for the data labeling process. ISO/TR 4804:2020 \cite{ISO-4804} specifies that a labeling quality report should be provided
during the development of the dataset. UL 4600 \cite{UL4600} suggests that perception shall map
sensor inputs to the perception ontology with acceptable performance and requires an
acceptably low incidence of mislabeled training and validation data. This standard also
requires arguments that machine learning training and testing datasets should have
an acceptable level of integrity and quality of data labels.
However, there are no standardized protocols for label review.

Labeling noise is seldom assessed in real-world settings.
But even in parts meant for testing and validation, the error rate in well-known datasets can
reach 10\% \cite{Northcutt2021PervasiveLE}.
This error rate may drastically reduce the maximum accuracy of DL algorithms.
Furthermore, with such a high rate of labeling errors, it follows that
algorithms shouldn't be expected to achieve the maximal accuracy possible.
While there are training methods that are designed to offset the impact 
of the labeling noise in detection problems, such methods usually 
are not able to directly assess the quality of the labels. 

In this paper, we introduce Confident Learning for Object Detection (CLOD).
We present an overview of CLOD in Fig.~\ref{fig:meme}.
This work builds on confident learning (CL) \cite{Northcutt2021ConfidentLE,Thyagarajan2023MultilabelCL},
one of the state-of-the-art techniques for detecting noisy labels in classification tasks.
We also extend the method for the evaluation of image recognition datasets from
\cite{Popowicz2022CombatingLN} toward the object detection task.
To the best of our knowledge, this is the first time the CL technique is applied to
object detection to find all types of label issues.
Our extensive assessment revealed that the proposed CLOD method can very effectively detect incorrect annotations. As such, this approach can be used in automated evaluation of the quality of object detection datasets.

The main contributions of this paper are the following:
\begin{enumerate}
    \item We introduce the CLOD algorithm based on confident learning. To the best of our knowledge,
it is the first model-agnostic method that focuses on the detection of erroneous labels in object detection datasets,
as existing methods focus mainly on offsetting the label noise in the training process.
\item  We provide an extensive empirical evaluation using two state-of-the-art
datasets: COCO and Waymo using well-known DNN architectures: Faster R-CNN, RetinaNet
and YOLO.
\item  We present examples from these two datasets that the proposed method
marked as suspicious, revealing that even commonly used datasets can contain erroneous annotations. 
\item We experimentally confirm that correction of the automatically identified noisy labels significantly improves the performance of the object detector in comparison to training on noisy data.
\end{enumerate}

% ------------------------------------------------------------------------------

\section{Related work}

Recently, a lot of research focused efforts on labeling noise.
The easiest way to assess a training dataset's quality is to thoroughly tag it
several times using multiple annotators.
The labels that each annotator assigned to each image may then be compared to spot errors.
\cite{LampertStumpf2016} introduced inter-annotator agreement measures
that may be used to gauge the degree of agreement
among a set number of annotators when assigning category labels.
However, because this approach requires annotating the dataset more than once, this
significantly raises preparation costs.

There are also fully automated methods dealing with noise in object detection datasets.
These studies, however, focus on implementing adaptive, noise-tolerant DL
algorithms rather than attempting to quantify the number of false labels
discovered.
Therefore, it is impossible to utilize these methods to check the correctness of training
datasets supplied by an external source. Such methods are only suitable for the training
of models on noisy datasets. We present an overview of these methods in the following subsections.

% ------------------------------------------------------------------------------

\subsection{Weighting noisy samples}

Reed et al. proposed a modification to the loss function \cite{Reed2015TrainingDN} that lowers the
impact of labels that might be inaccurate. 
The authors claim that this technique also permits the use of datasets whose
labels are not known in their entirety. The authors tested this method in both image recognition and object detection.

Goldberger and Ben-Reuven developed the S-Model architecture
\cite{Goldberger2017TrainingDN} by adding another softmax layer to the DNN
model for classification tasks.
The authors assume a particular distribution of noise and use the expectation-maximization method.

Different weights are given to the input images by the MentorNet classifier network
presented by Jiang et al. \cite{Jiang2018MentorNetLD}.
In successive iterations of the algorithm, it gives lower weights to the images whose labels are assigned with
low probability. As a result, such images effectively stop
contributing to the learning process.
The iterative cross-validation approach \cite{Chen2019UnderstandingAU} described
in a more recent work identifies which photos from the training classification dataset
should be deleted in the following iteration. However, the removed samples
are not corrected after detection. The authors demonstrate how a DNN 
trained with such recurrent pruning performs better.

% ------------------------------------------------------------------------------

\subsection{Ignoring possibly-incorrect labels}

\cite{Han2018CoteachingRT} introduced co-teaching for classification.
This approach utilizes two DNNs to determine whether an image has
been erroneously tagged by sharing training datasets. From each mini-batch, the algorithm uses only a fraction
of examples with the lowest loss on one network to train the other network.
% For a label to be deemed incorrect, it must be decided by both collaborating
% DNNs.
In \cite{Chadwick2019TrainingOD}, authors introduce a modification to co-teaching
\cite{Han2018CoteachingRT} to treat objects independently within an image.
Standard co-teaching forces one to ignore the whole image as being noisy.
In their solution, the authors only select affected annotations in the image.
The idea is not to ignore entire images because of a few incorrect annotations.

Another work \cite{Mao2021NoisyAR} proposes an additional layer in the
Faster R-CNN network to separate errors in the location of the bounding box
from the errors in the label.
The authors showed that this approach directly separates the erroneous areas from the
correct ones.
For this, they use the Cross-Iteration Noise Judgment technique.
This method assumes that the percent of noisy labels is constant, thus discarding a predefined number of images.
Such an assumption cannot be made a priori in real data.

Unfortunately, none of the methods described above can point out if the images were
properly tagged or whether they had any atypical content, making the detection task
more difficult and degrading the DNN's effectiveness.
The goal of these techniques is to elevate the final accuracy of DNNs either by
rejecting training samples or lowering their weight in the training process.
Even if they select image labels as potentially invalid, they do that 
for a given constant fraction of examples given as parameters,
while this fraction may vary between datasets \cite{Northcutt2021PervasiveLE} 
and should be treated as unknown.

\subsection{Automatic correction of annotations}

A two-step algorithm was proposed in \cite{Liu2022TowardsRA} for noise
mitigation in object detection datasets.
The authors suggest performing class-agnostic bounding box correction, which
optimizes the noisy ground-truth boxes regardless of class labels.
Then, in the second step, labels are corrected. Both steps adjust all
annotations without selecting the incorrect ones.
Eventually, this method gradually modifies the training set toward better outcomes of
trained DNN models.

In \cite{Wang2022NoisyLocationAnnotations}, Wang et al. adopted teacher-student
networks to mitigate label noise. The method considers network predictions
as statistical measurements of the correct bounding boxes. A teacher
network is trained on noisy labels, and its outputs are used to estimate
the correct bounding boxes to train the student network.

% ------------------------------------------------------------------------------

\subsection{Confident learning-based methods}

In computer vision, the detection of erroneous annotations as the main problem to be solved has been addressed so far only for the image recognition problem.
The confident learning (CL) introduced in \cite{Northcutt2021ConfidentLE} is a promising approach.
The method uses N-fold cross-validation, as in \cite{Chen2019UnderstandingAU},
but only once rather than repeatedly.
CL chooses each class's probability cutoffs separately.
The cutoffs decide whether a particular sample can be confidently categorized.
If an image is incorrectly yet confidently categorized, it is viewed as incorrectly labeled. Such examples are deleted from the dataset to improve performance.
A DNN model is then created using the newly filtered data.
The designers of the CL technique 
used the ResNet-50 architecture  to
assess the approach's efficacy using a dataset artificially affected by relatively high (20--70\%) mislabel
probability.
Unfortunately, the authors presented no records of how many incorrect images CL eliminated
from the dataset.
The method, like in the majority of earlier works presented in this section,
focuses on improving the final DNN model's classification performance.
This approach results in larger improvements than all the
other techniques examined here. 
A recent work \cite{Thyagarajan2023MultilabelCL} adapts CL to
multi-label classification tasks. However, it also does not report how many 
noisy labels were detected. Instead, it only measures the improvement of 
the classifier's performance.

\cite{Northcutt2021PervasiveLE} presents examples of how known image
datasets have been analyzed using the CL approach.
The researchers chose suspicious images and manually checked them to find incorrect labeling.
They found that the state-of-the-art datasets for image recognition contain between 0.15 and 10\% images with incorrect labels.
This is a noteworthy discovery, given that the examined reference databases should be devoid of such errors.
Such a significant fraction of wrong labels makes questionable some popular competitions among classifier algorithms that frequently reach accuracy above 99\%.
Moreover, the number of verified noisy labels in this investigation was only a small part of the initial CL predictions.
It suggests that further balancing accuracy versus recall can enhance the performance of
the label noise detection with CL.

In \cite{Popowicz2022CombatingLN}, Popowicz et. al utilized CL with multiple
DNN models to increase the ability to detect noisy labels in image classification datasets.
They introduced a parameter, which adjusts the sensitivity of noise detection and proposed using multiple DNN models to increase the
detection reliability. This paper was the first work concentrating not on the accuracy of DNN trained on a cleaned dataset, but on the actual detecting capability of the CL technique.

Finally, Tkachenko et. al \cite{objectlab} presented a method for calculating
box quality scores in object detection. This is achieved by weighting three
kinds of scores: \textit{overlook} score (object not annotated), \textit{badloc}
score (imprecise bounding box) and \textit{swap} score (wrong class label).
This method does not consider spurious bounding boxes.
Authors evaluate ObjectLab and other methods by gauging their performance on
a \textit{per image} basis, i.e. they calculated the image annotation quality by
aggregating quality scores of all corresponding boxes.
We compared ObjectLab to CLOD by considering the quality of individual bounding boxes
in sec. \ref{sec:experiments}.

% ------------------------------------------------------------------------------

\section{Confident learning}

Confident learning (CL) points out suspicious labels by assuming class-conditional
noise and estimating the joint probability distribution between noisy and 
correct labels. This section summarizes the multi-label CL method,
as introduced in \cite{Thyagarajan2023MultilabelCL}.

For each possible label $m$, each input example $x_i$
in the noisy dataset $\mathbf{X}$ is assigned a noisy binary label $y_i^m$. 
It sets $y_i^m = 1$ if $x_i$ has the noisy label $m$, else $y_i^m = 0$.
Using model-predicted label probabilities $p_i^m$, the \emph{self-confidence}
score $s_i^m$ is calculated as
\begin{equation}
    s_i^m = y_i^m p_i^m + (1 - y_i^m) (1 - p_i^m)
\end{equation}
for each example $i$ and label $m \in 1, \ldots M$.

Next, the algorithm aggregates self-confidence scores using one of the proposed
pooling methods. The authors report that the exponential moving average works 
well, when compared to other considered pooling methods, thus we use it in this work. For $\check{s}_i^1 \geq \ldots \geq \check{s}_M^1$
denoting sorted values $s_i^m$ the aggregated score $s_i^*$ is defined as
\begin{equation}
    s_i^*=S_i^M \text{ where } S_i^t = \begin{cases}
    \check{s}_i^1 & \text{if } t = 1 \\
    \alpha \check{s}_i^t + (1-\alpha) S_i^{t - 1} & \text{if } t > 1
    \end{cases}
\end{equation}
where $\alpha$ is a parameter. The authors of multi-label CL show that a value of $\alpha=0.8$ gives good results.
Such a formulation makes the aggregated label quality score $s_i^*$ most heavily influenced
by the lowest values of $s_i^m$.

Similarly to the original CL method \cite{Northcutt2021ConfidentLE}, 
the suspected labels may be selected by using the rank-and-prune
approach, i.e., by selecting a given fraction of examples with the lowest values
of the aggregated self-confidence score $s_i^*$.

% ------------------------------------------------------------------------------

\section{Confident Learning for Object Detection (CLOD)}
\label{sec:cl}

Existing CL methods cannot directly analyze an object detection dataset.
Assigning a set of labels to an example (image) does not take into account
bounding boxes.
We propose a method for applying CL for the object detection task to generate box quality scores and correction suggestions.

For each image $x$ in the noisy dataset $\mathbf{X}$ an object detection
model is used to obtain labeled bounding boxes through inference.
As the samples are later processed by the CL algorithm, the predictions
should be out-of-sample to avoid bias caused by overfitting.
Original and model-predicted (MP) boxes are then clustered with the following
rules. 1.~The clustering is based on a distance threshold. 2.~The clustering distance between two boxes $a$ and $b$ is calculated as $1 - IoU(a,b)$, where $IoU$ is intersection over union. 3.~Each cluster must contain only boxes related to the same image.

\begin{figure}[htb]
    \centering
    \includegraphics[width=0.6\linewidth, trim={0 32 0 37}, clip]{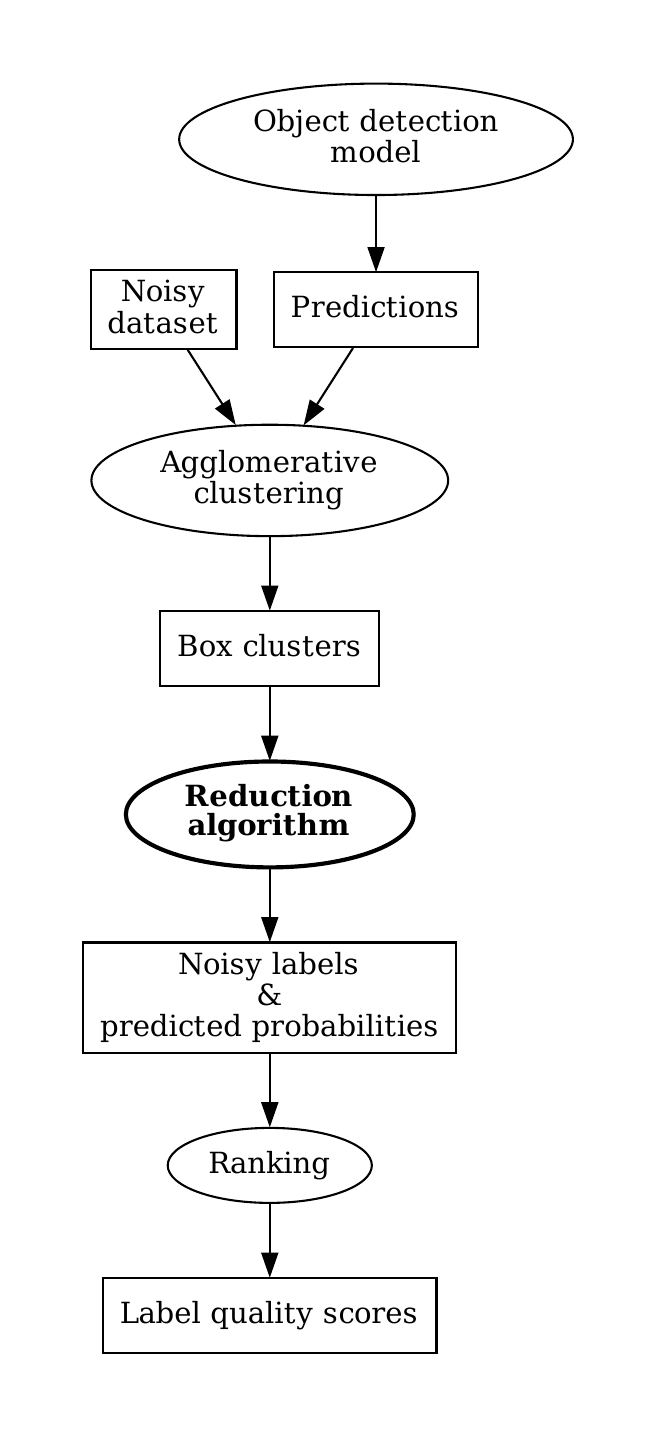}
    \caption{Data flow diagram for CLOD.}
    \label{fig:flow}
\end{figure}

For clustering of the bounding boxes, we propose agglomerative clustering
with single linkage \cite{HAC}.
We then perform reduction, processing each such cluster separately.
Fig.~\ref{fig:flow} provides an overview of the method and Alg.~\ref{alg:reduction}
describes the reduction procedure.

From each cluster
$\mathbf{C}_k^\mathrm{X} \bigcup \mathbf{C}_k^{\mathrm{\hat{P}}}$,
where $\mathbf{C}_k^\mathrm{X}$ are original annotations in $k$-th
cluster and $\mathbf{C}_k^{\mathrm{\hat{P}}}$ are model predictions in $k$-th cluster, we calculate one row $\mathbf{Y}'_k$ of the new binary
label matrix $\mathbf{Y}'$ and one row $\mathbf{\hat{P}'}_k$ of the new predicted
probabilities matrix $\mathbf{\hat{P}'}$.
For $M$ denoting the number of possible labels
an element $\mathbf{Y}'_{k,m}; m = 1, 2, \cdots M$ is equal to $1$ if there is a box with label $m$ in
$\mathbf{C}_k^\mathrm{X}$ and $0$ otherwise.
If there are no elements set to $1$ in this row, then the \textit{background} label $\mathbf{Y}'_{k,M+1}$ is set to $1$. Otherwise, it is set to $0$.
It means that if the current cluster does not contain any boxes from the
original noisy dataset, the cluster represents \textit{background}.

Then elements of the $\mathbf{\hat{P}'}_k$ row are calculated.
$\mathbf{\hat{P}'}_{k,m}$ is equal to the maximum score of model-predicted
boxes of label $m$ in current cluster, or $0$ if there are no such boxes.
Same as in the case of $\mathbf{Y}'_{k,M+1}$, $\mathbf{\hat{P}'}_{k,M+1}$ is
equal to $1$ if the row contains only zeros and $0$ otherwise.
This element represents the model-predicted score of \textit{background}.

\begin{algorithm}
    \caption{The reduction algorithm.}\label{alg:reduction}
    \KwData{$\mathbf{C}_k^\mathrm{X}$, $\mathbf{C}_k^{\mathrm{\hat{P}}}$}
    \KwResult{$\mathbf{Y}'_k$, $\mathbf{\hat{P}'}_k$}
    \For{$m$ in $1$..$M$}{
      $\mathbf{Y}'_{k,m} = 1$ if $m$ is in labels of $\mathbf{C}_k^\mathrm{X}$ else $0$\;
    }
    $\mathbf{Y}'_{k,M+1} = 1$ if $\sum_{m=1}^{M}\mathbf{Y}'_{k,m} = 0$ else $0$\;
    \For{$m$ in $1$..$M$}{
      $\mathbf{\hat{P}'}_{k,m}
        = \mathrm{max}($scores of $\mathbf{\hat{P}'}_k$ where label is $m)$\;
    }
    $\mathbf{\hat{P}'}_{k,M+1} = 1$ if $\sum_{m=1}^{M}\mathbf{\hat{P}'}_{k,m} = 0$ else 0\;
    \vspace{0.5cm}
\end{algorithm}

This way, the whole dataset is processed, cluster by cluster,
and the final matrices $\mathbf{Y}'$ and $\mathbf{\hat{P}'}$ are fed to the
CL algorithm. The results are the aggregated $s^*_k$ scores for each row $k$ of the $\mathbf{Y}'$ and $\mathbf{\hat{P}'}$ matrices.
We assign the label quality score $s^*_k$ to each annotation in the $\mathbf{C}_k^\mathrm{X}$ set. If there are no annotations in $\mathbf{C}_k^\mathrm{X}$, we return the model-predicted annotations $\mathbf{C}_k^\mathrm{\hat{P}}$ as possible missing annotations with the quality score of $s^*_k$.

\begin{figure}[htb]
    \centering
    \includegraphics[width=\linewidth]{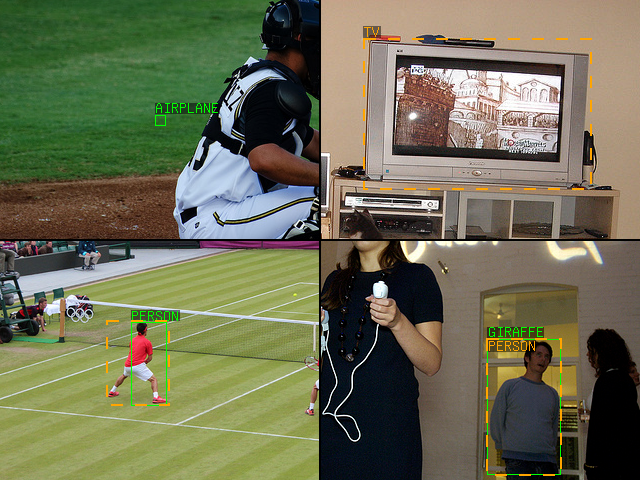}\\
    %\vspace*{10pt}
    \caption{Examples of spurious (top left), missing (top right), mislocated (bottom left), and mislabeled boxes (bottom right). Green: original boxes, orange: model predictions.}
    \label{fig:issue-ex}
\end{figure}

Similarly to both original and multi-label CL algorithms, a rank-and-prune (removing a percentage of images with the lowest annotation confidence scores) may be used to select the annotations that are the most likely to be erroneous.
We can also use the results of clustering to guess what type of error we
might be dealing with.
Considering each cluster separately: 
\begin{enumerate}
    \item If there are only model predictions in the cluster, it indicates a possibly missing annotation (false negative),
    \item If there are no model predictions in the cluster, it indicates a possibly spurious annotation (false positive),
    \item The presence of annotations and predictions with different labels indicates a possibly mislabeled annotation,
    \item The presence of annotations with the same labels indicates a possibly mislocated annotation.
\end{enumerate}
See Fig.~\ref{fig:issue-ex} for examples of each issue.

This approach can also be extended to formulating suggestions for correcting
annotations.
In the case of missing, mislabeled, and mislocated annotations, the model
predictions that are part of the cluster can be considered suggestions.

% ------------------------------------------------------------------------------

\section{Experiments}
\label{sec:noisydata}

We conducted five experiments to examine the effectiveness of CLOD.
The experiments are varied to assess different aspects of this method.
We used multiple datasets (see Sec.~\ref{sec:datasets}) and model architectures
(see Sec.~\ref{sec:models}).
We also considered both artificial noise (see Sec.~\ref{sec:noise}) and real-world noisy labels present in the analyzed datasets.

\subsection{Datasets}
\label{sec:datasets}

We used Common Objects in Context (COCO) 2017 and Waymo Open Dataset v1.4.0.
This choice was dictated by the prominence of COCO in the evaluation of object
detection models and the popularity of Waymo in the autonomous vehicle field,
which is the main interest of our research team.
Please note that the original Waymo dataset was published in the form of an annotated multi-camera video
sequences.
From each sequence, we extracted 10 front camera frames, equally spaced in time, to work with
still images.
We also used the Sama-Coco dataset \cite{sama-coco} as a human-revised variant of COCO.
It is worth noticing that the authors of Sama-Coco added more than 200k object annotations to the original dataset.
That means that in their opinion 25\% of the annotations are missing from the original dataset.
Nevertheless, many of them refer to very small objects and it strongly depends on the labeling policy whether they should be annotated or not.

Additionally, we created a variant of the Waymo dataset in which, due to resource constraints, 12\% of the dataset images were manually reviewed by our team.
Please refer to Tab.~\ref{tab:datasets} for details about the datasets.

\begin{table}[h]
  \centering
  \caption{Datasets.}
  \begin{tabular}{@{}lrr@{}}
    \toprule
    Dataset & Images & Instances \\
    \midrule
    Common Objects in Context 2017 & 123,287 & 886,284 \\
    \hspace{0.25cm} training & 118,287 & 860,001 \\
    \hspace{0.25cm} validation & 5,000 & 36,781 \\
    Sama-Coco & 123,287 & 1,115,464 \\
    \hspace{0.25cm} training & 118,287 & 1,067,995 \\
    \hspace{0.25cm} validation & 5,000 & 47,469 \\
    \midrule
    partial Waymo Open Dataset v1.4.0 & 10,000 & 230,479 \\
    \hspace{0.25cm} training & 7,980 & 185,470 \\
    \hspace{0.25cm} validation & 2,020 & 45,009 \\
    revised p. Waymo Open Dataset v1.4.0 & 10,000 & 229,426 \\
    \hspace{0.25cm} training & 7,980 & 184,830 \\
    \hspace{0.25cm} validation & 2,020 & 44,596 \\
  \end{tabular}
  
  \label{tab:datasets}
\end{table}

% ------------------------------------------------------------------------------

\subsection{DNN architectures}
\label{sec:models}

We selected three DNN architectures for the experiments.
They were Faster R-CNN \cite{Ren2015FasterRT}, RetinaNet \cite{Lin2017FocalLf}
and YOLO \cite{Redmon2018YAI}.
The choice was dictated by the fact that these are state-of-the-art solutions
for object detection.
They provide enough variation to evaluate the algorithm's independence on the
type of network used to generate predictions.
We provide training parameters for these models in Tab.~\ref{tab:model-params}.

\begin{table}[ht]
\centering
\caption{Training parameters}
\begin{tabular}{@{}lccc@{}}
\toprule
Parameter & Faster R-CNN & RetinaNet & YOLO \\
\midrule
Epochs & 12 & 12 & 273 \\
LR & 0.02 & 0.01 & 0.001 \\
Batch size & 2 / worker & 2 / worker & 8 / worker \\
\midrule
Transforms & \multicolumn{2}{c}{\makecell{resize to 1333x800\\(COCO only), \\ random flip, \\ normalization}} & \makecell{random crop, \\ random resize\\320x320 or 608x608\\(COCO only), \\ random flip, \\ photometric distortion, \\ normalization} \\
\midrule
Optimizer & \multicolumn{3}{c}{Stochastic gradient descent} \\
\end{tabular}
\label{tab:model-params}
\end{table}

% ------------------------------------------------------------------------------

\subsection{Artificial noise}
\label{sec:noise}

In some of the experiments, we used artificial label noise.
It can be added on-demand and with granular control over its parameters.
Moreover, the knowledge of which bounding boxes were disturbed enabled us to
thoroughly evaluate our method.
To calculate the exact number of true and false positives and negatives we
would have to assume that the original dataset was flawless.
However, since we are adding high-amplitude noise to already noisy examples
only false positive counts should be slightly biased.

We selected five noise types that imitate real-world annotation noise
\cite{Chadwick2019TrainingOD,Mao2021NoisyAR,Li2020TowardsNO}.
\textit{Uniform label noise} was created by changing the labels of bounding
boxes to ones selected randomly from the set of labels, with equal
probability.
\textit{Location noise} was a disturbance to the location of boxes by
selecting a random angle and moving the bounding boxes along this direction
by a vector of length proportional to the box's dimensions.
\textit{Scale noise} was growing or shrinking bounding boxes by a fixed factor.
\textit{Spurious boxes} of random dimensions and labels were added to the
dataset
and \textit{missing noise} was applied by removing random boxes.

\subsection{Environment}
\label{sec:env}

We performed all of the experiments on a server with an AMD EPYC 7313 16-core processor, RTX A5000 graphics processing unit, and 60 GB of memory.
The server was running Ubuntu 22.04, MMDetection 2.28.2, and PyTorch 1.12.1 with CUDA 11.3.

\subsection{Experiments and results}
\label{sec:experiments}

\begin{figure}[htb]
    \centering
    \includegraphics[width=\linewidth]{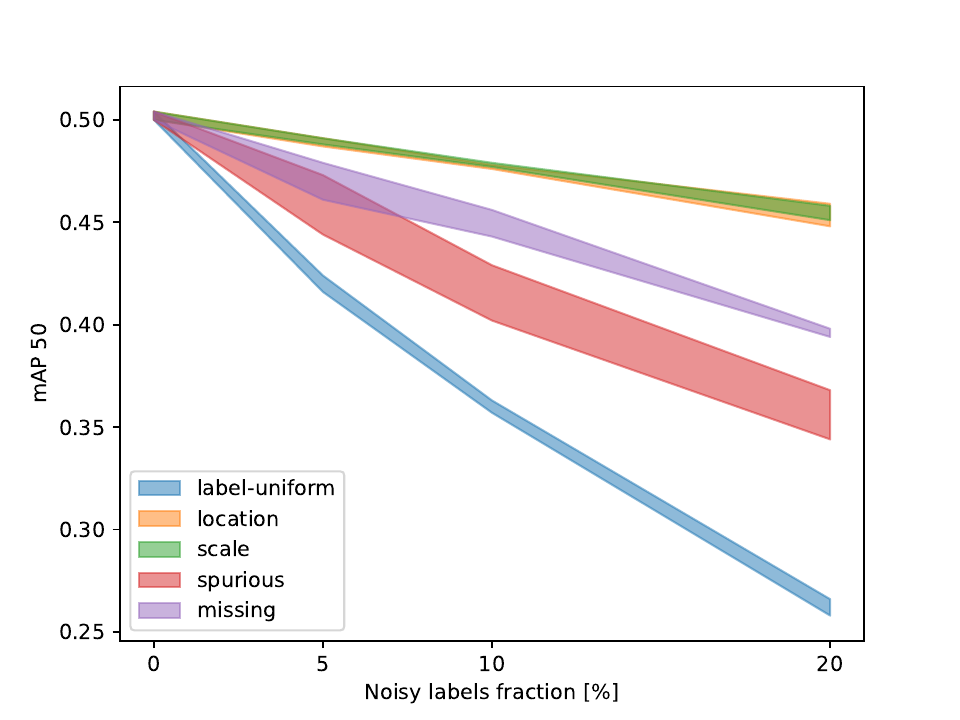}
    \caption{Influence of different types of noise in the
      training dataset on model quality. Bands represent the stretch of values across different model architectures.}
    \label{fig:noisy-map}
\end{figure}

{\bf \noindent Noise impact.} The purpose of the first experiment was to measure the impact of the annotation noise on the quality of models.
We applied artificial noise to increasing fractions of incorrect labels of the COCO training
and validation datasets.
We trained models and evaluated them using mean average precision at IoU = 0.5.
The relationship between evaluation results and the fraction of noisy boxes
was recorded and presented in Fig. \ref{fig:noisy-map}.
\todo[inline, author=kuba, disable]{Why 0.5?}

The results show that every type of training dataset noise significantly decreases
the quality of trained models.
This effect is consistently greater with larger amounts of noisy boxes.
Uniform label noise was the most detrimental to the quality of the models,
followed by spurious boxes and missing boxes.
Location and scale noise (with the magnitude of 25\% of box dimensions) were the
least influential.

\begin{figure}[htb]
    \centering
    \includegraphics[width=\linewidth]{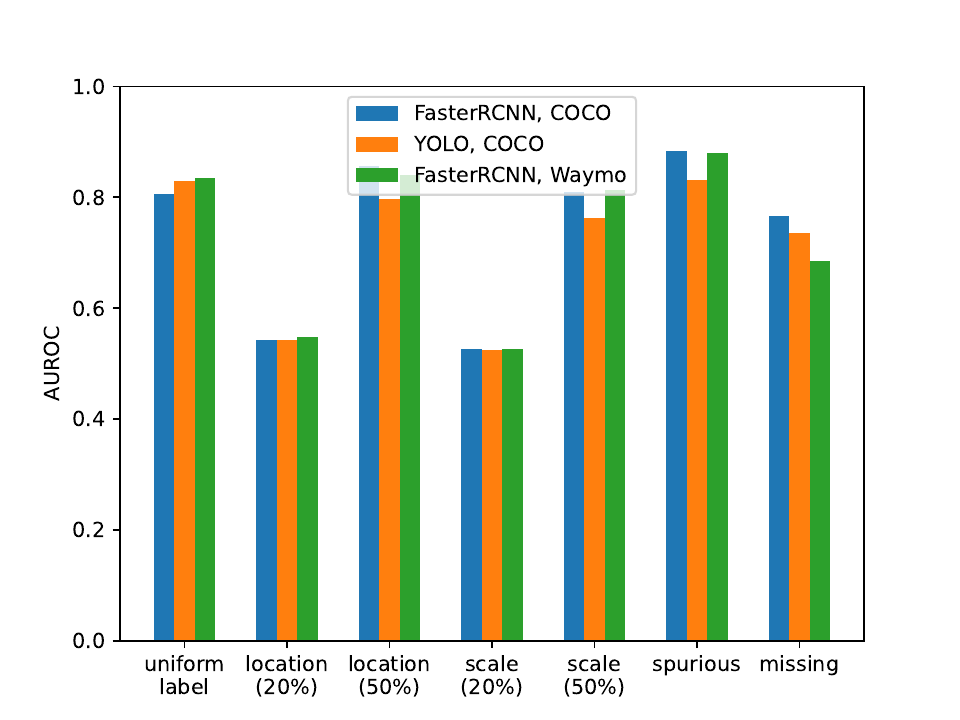}
      \caption{Effectiveness of the CLOD method in detecting different
        types of artificial noise. IoU clustering threshold: 0.5. Median from 3 experiments.}
    \label{fig:cl-eval}
\end{figure}

{\bf \noindent CLOD effectiveness.} Next, we evaluated the CLOD algorithm using the same artificial label noise.
We trained models on training parts of selected datasets.
Then, we added noise to 20\% of examples from the corresponding validation datasets.
CLOD was used to
find the disturbed examples in these validation datasets.
We calculated the area under the receiver operating characteristic curve (AUROC)
against the label quality threshold for different types of artificial noise.
We present the final results in Fig.~\ref{fig:cl-eval}.
\todo[inline, author=kuba, disable]{Why 20\%?}

The proposed method is best at detecting spurious bounding boxes but 
it detects label noise almost as well.
It did not, however, succeed in pointing out low amplitude location and scale
noise, showing performance similar to a random classifier.
However, it is consistent across different datasets and model architectures.

\begin{figure}[htb]
    \centering
    \includegraphics[width=\linewidth]{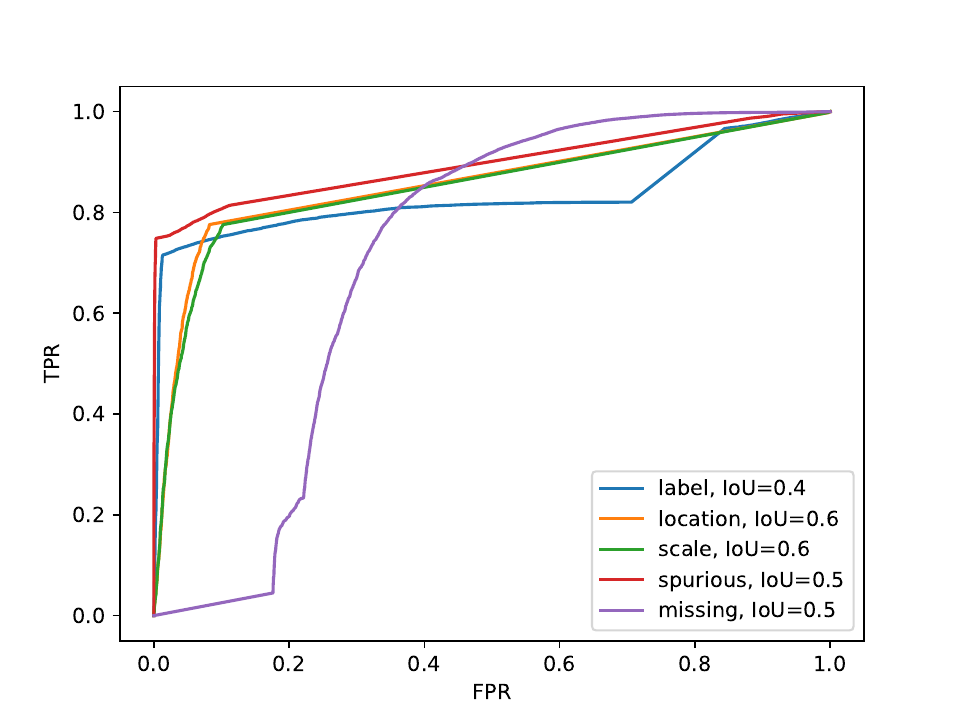}
      \caption{A few examples of CLOD ROC curves.}
    \label{fig:roc}
\end{figure}

A glance at the ROC curves (Fig. \ref{fig:roc}) reveals that CLOD can detect nearly all types of noise well.
Considering all but missing examples, the method points out nearly 80\% of noisy boxes with
a false positive rate under 0.1.
For the missing examples, however, it seems that to achieve the same, one needs to accept a 0.3 false positive rate.

\begin{figure}[htb]
    \centering
    \includegraphics[width=\linewidth]{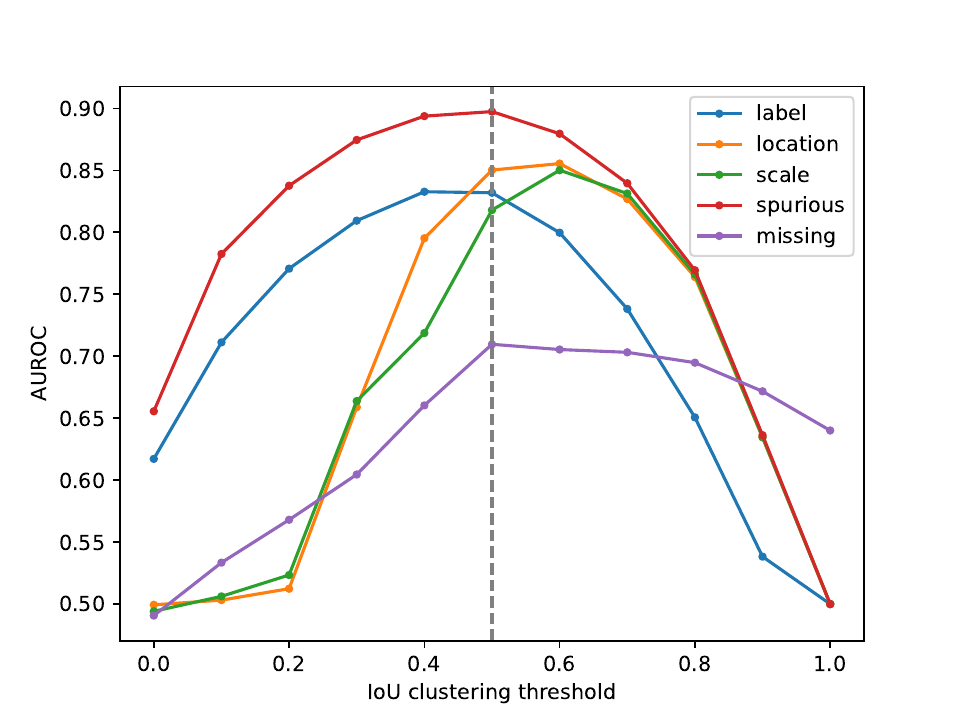}
      \caption{AUROC of the CLOD method with respect to the clustering threshold.
        COCO dataset, Faster R-CNN.}
    \label{fig:coco-frcnn}
\end{figure}

We chose the clustering threshold based on earlier experiments that
compared results for a range of thresholds.
We included the results in Fig.~\ref{fig:coco-frcnn}.
Highest-scoring thresholds were 0.4, 0.5, and 0.6, which makes 0.5 a good default threshold.

{\bf \noindent CLOD vs state-of-the-art.}
We also compared CLOD to ObjectLab \cite{objectlab}.
Artificial noise (see sec. \ref{sec:noise}) was added to 25% of examples in the COCO and Waymo validation datasets.
Then, we ran CLOD and ObjectLab to detect the noisy boxes.
We evaluated the performance of the methods by calculating AUROC for every combination of noise types and datasets.
The results are presented in Tab.~\ref{tab:comparison}.

\begin{table}
  \centering
  \caption{AUROC values for different noise types for CLOD and ObjectLab. The best parameters were determined for all methods by a grid search. \todo[inline, author=kuba]{Grid search over? maybe move info about parameters to text.}}
  \label{tab:comparison}
  \begin{tabular}{@{}llrr@{}}
    \toprule
    Noise & Dataset & CLOD & ObjectLab \\
    \midrule
    label & COCO & 0.833 & \textbf{0.851} \\
    label & Waymo & 0.746 & \textbf{0.854} \\
    \midrule
    location & COCO & \textbf{0.855} & 0.835 \\
    location & Waymo & 0.845 & \textbf{0.851} \\
    \midrule
    scale & COCO & \textbf{0.850} & 0.822 \\
    scale & Waymo & 0.825 & \textbf{0.828} \\
    \midrule
    spurious & COCO & \textbf{0.897} & N/A \\
    spurious & Waymo & \textbf{0.967} & N/A \\
    \midrule
    missing & COCO & \textbf{0.710} & 0.524 \\
    missing & Waymo & \textbf{0.705} & 0.564 \\
    \bottomrule
  \end{tabular}
\end{table}

CLOD scored, on average, around 30\% higher than ObjectLab in detecting missing boxes.
It also performed well in detecting spurious examples, which is a case that ObjectLab does not consider.
ObjectLab however, scored 14\% higher when considering mislabeled boxes in the Waymo dataset.
In other cases, the two methods achieved similar scores.
\todo[inline, author=kuba]{Why is ObjectLab better in some cases and CLOD in other? Maybe we could try to explain the results.}

{\bf \noindent Dataset variants.}
The fourth experiment was to show the performance of CLOD in real-world settings.
We compared model quality scores between multiple variants of COCO and Waymo
datasets while treating the respective original datasets as a baseline.
We used human-revised variants of both COCO and Waymo datasets to further assess the importance of data quality.

Creating the other two variants was the main part of this experiment.
We wanted to gauge how accepting CLOD suggestions would affect the overall
quality of the dataset.
The \textit{suggestions} datasets are the original datasets with the best 20\% of CLOD suggestions
applied.
The \textit{selected} datasets were created by matching suggested bounding boxes to
those from the human-revised variants based on IoU (>= 0.5).
\todo[inline, author=kuba, disable]{Why 20\%?}

Both training and validation datasets were corrected because both can contain errors.
Additionally, correcting the dataset alters the distribution of labels and sizes of bounding
boxes.
For example, if numerous small boxes were added to the training dataset but not to the validation dataset,
the model would not be evaluated fairly.

We found counterparts for the best 20\% of suggestions and applied them to the original dataset.
The purpose of creating this dataset was to check what effect accepting \textit{good} human-provided modifications and rejecting \textit{bad} ones has on the quality of the datasets.
This is under the assumption that CLOD rates annotations perfectly.
We provide the summary of the mentioned datasets in Tab.~\ref{tab:datasets2}.
We compared the mean average precision (mAP) of models trained and evaluated using described variants and recorded the values into Tab.~\ref{tab:variants}.

\begin{table}[h]
  \centering
  \caption{Datasets created using CLOD.}
  \label{tab:datasets2}
  \begin{tabular}{@{}lrr@{}}
    \toprule
    Dataset & Images & Instances \\
    \midrule
    COCO Suggestions & 123,287 & 4,085,931 \\
    \hspace{0.25cm} training & 118,287 & 3,924,926 \\
    \hspace{0.25cm} validation & 5,000 & 161,005 \\
    COCO Selected & 123,287 & 996495 \\
    \hspace{0.25cm} training & 118,287 & 955,077 \\
    \hspace{0.25cm} validation & 5,000 & 41,418 \\
    \midrule
    Waymo Suggestions & 10,000 & 178,964 \\
    \hspace{0.25cm} training & 7,980 & 144,713 \\
    \hspace{0.25cm} validation & 2,020 & 34,251 \\
    Waymo Selected & 10,000 & 229,426 \\
    \hspace{0.25cm} training & 7,980 & 184,830 \\
    \hspace{0.25cm} validation & 2,020 & 44,596 \\
    \bottomrule
  \end{tabular}
\end{table}

\begin{table}
    \scriptsize
    \caption{Faster R-CNN model quality scores, trained using datasets created using CLOD.}
    \label{tab:variants}
    \centering
    \begin{tabular}{lcccccc}
        \toprule
        Dataset & $\mathrm{mAP}$ & $\mathrm{mAP_{50}}$ & $\mathrm{mAP_{75}}$  \\
        \midrule
        \textbf{COCO} & & & \\
        Original & 0.316 & 0.501 & 0.342  \\
        Revised & 0.318 & 0.492 & 0.347  \\
        Suggest. & \textbf{0.366} & \textbf{0.610} & \textbf{0.379} \\
        Selected & 0.356 & 0.572 & 0.379  \\
        \midrule
        \textbf{Waymo} & & &  \\
        Original & 0.310 & 0.548 & 0.316  \\
        Revised & 0.237 & 0.415 & 0.242  \\
        Suggest. & \textbf{0.452} & \textbf{0.774} & \textbf{0.471}  \\
        Selected & 0.315 & 0.554 & 0.316  \\
        \bottomrule
    \end{tabular}
\end{table}

\begin{table}
    \scriptsize
    \caption{Faster R-CNN model quality scores for various object sizes.}
    \label{tab:variants2}
    \centering
    \begin{tabular}{lcccccc}
        \toprule
        Dataset &  $\mathrm{mAP_{S}}$ & $\mathrm{mAP_{M}}$ & $\mathrm{mAP_{L}}$ \\
        \midrule
        \textbf{COCO}  & & & \\
        Original &  0.181 & 0.344 & 0.400 \\
        Revised &  0.194 & 0.351 & \textbf{0.420} \\
        Suggest.  & \textbf{0.325} & \textbf{0.383} & 0.359 \\
        Selected  & 0.252 & 0.367 & 0.414 \\
        \bottomrule
    \end{tabular}
\end{table}

Results show that the revised Sama-Coco dataset scores very similarly to the
original COCO dataset.
This is not the case with the revised Waymo, which scored much lower.
The corrections suggested by CLOD scored the highest, and selected datasets
came out in between.
Additionally, we calculated the metrics $\mathrm{mAP_{S}}$, $\mathrm{mAP_{M}}$, $\mathrm{mAP_{L}}$
for small, medium, and large objects based on the split provided by the authors of the COCO dataset.
We present the results in Fig. \ref{tab:variants2}.

{\bf Manual inspection. }The last experiment was to run CLOD on the original datasets
and view annotations that were deemed suspicious.
Some examples are presented and commented on below.
We also took note of the annotation rating time and peak memory usage.
We present these results in tab.~\ref{tab:timemem}.
On average, CLOD processed approximately 4000 annotations per second.
It used approximately 4 KiB of memory per annotation on average.

\begin{table}[h]
  \centering
  \caption{Time and memory usage during label rating.}
  \label{tab:timemem}
  \begin{tabular}{@{}lrrr@{}}
    \toprule
    Dataset & Predictions & Time [s] & Memory [GiB] \\
    \midrule
    COCO & 3941531 & 1027.97 & 16.95 \\
    Sama-Coco & 3888731 & 1059.9 & 17.35 \\
    partial Waymo & 621403 & 61.75 & 1.74 \\
    \bottomrule
  \end{tabular}
\end{table}

Additionally, having analyzed the distribution of quality scores we concluded that,
in the COCO dataset, 25144 annotations (2.8\% of the dataset, quality cutoff: 3.9\%) should be
manually reviewed. When it comes to the Waymo dataset, only 878 annotations
(0.4\% of the dataset) fell below the 3.9\% quality threshold.
This threshold was picked by determining the point above which there is a significant increase
in the number of boxes with quality scores below it.

The first category of suspicious examples found in the COCO dataset represents groups of objects in a single box (Fig.~\ref{fig:examples-coco-groups}.
Each object should be annotated separately.
As the first image shows, such cases are not limited to groups of people but also objects like umbrellas.

% Some examples are commented out to conserve space.

\renewcommand\imw{0.73\linewidth}
\begin{figure}[phtb]
    \centering
    \includegraphics[width=\imw]{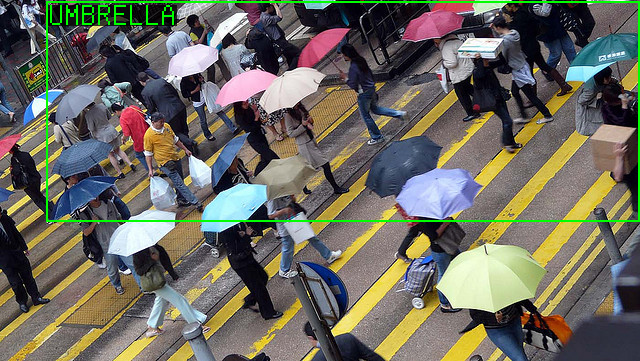}\\
    %\vspace*{10pt}
    %\includegraphics[width=\imw]{img/coco-sup/99.6a-900100072795-crowd.png}\\
    \vspace*{10pt}
    \includegraphics[width=\imw]{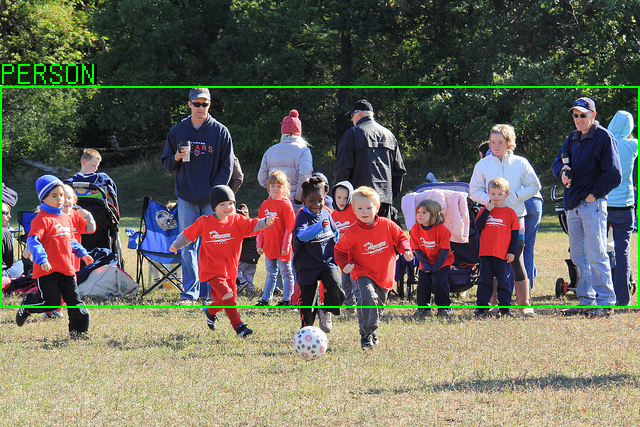}\\
    \vspace*{10pt}
    \includegraphics[width=\imw]{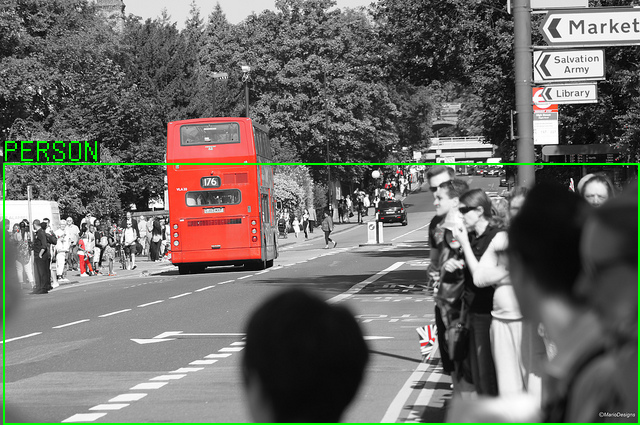}\\
    \vspace*{10pt}
    \includegraphics[width=\imw]{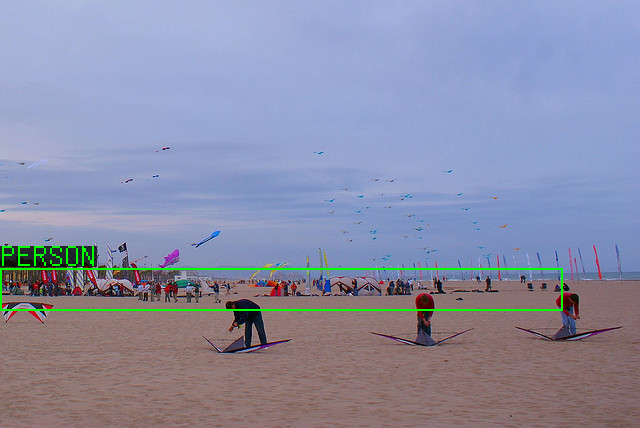}\\
    \vspace*{10pt}
    \includegraphics[width=\imw]{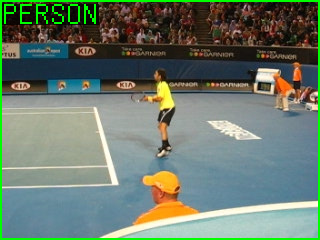}\\
    %\vspace*{10pt}
    %\includegraphics[width=\imw]{img/coco/whole-2.png}\\
    \caption{Groups of objects labeled as one object, COCO.}
    \label{fig:examples-coco-groups}
\end{figure}

The second category of boxes are those that label images of people as \textit{person}.
There are three examples in Fig.~\ref{fig:examples-coco-images}.
Two of them are photographs and one is a sculpture.
These boxes probably should not have been included in the dataset.

\renewcommand\imw{0.75\linewidth}
\begin{figure}[phtb]
    \centering
    %\includegraphics[width=\imw]{img/coco-sup/1.1a-000001744236-image.png}\\
    %\vspace*{10pt}
    \includegraphics[width=\imw]{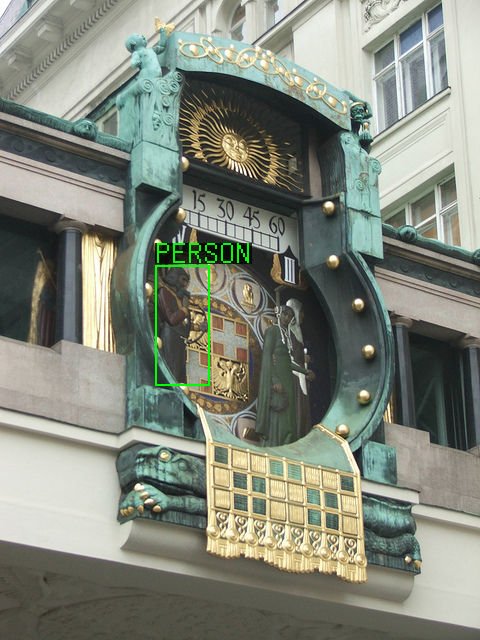}\\
    \vspace*{10pt}
    \includegraphics[width=\imw]{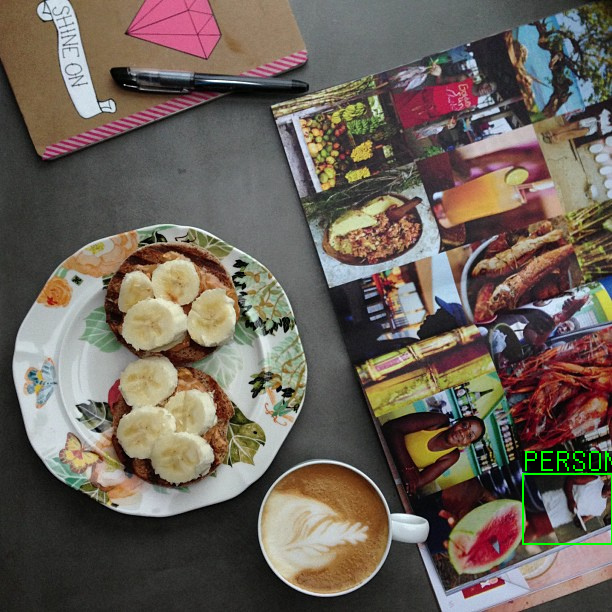}\\
    \vspace*{10pt}
    \includegraphics[width=\imw]{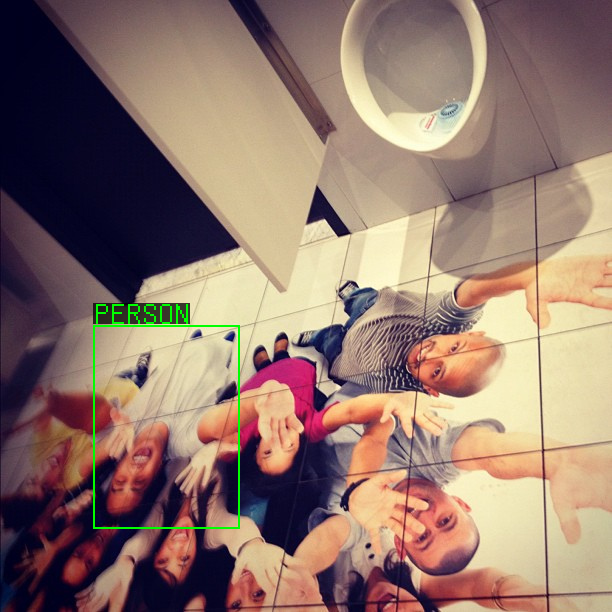}\\
    \caption{Images of people labeled as \textit{person}, COCO.}
    \label{fig:examples-coco-images}
\end{figure}

The next category is spurious boxes (Fig.~\ref{fig:examples-coco-spurious}).
These do not contain specified objects or the presence of such
objects is ambiguous.

\renewcommand\imw{0.76\linewidth}
\begin{figure}[phtb]
    \centering
    \includegraphics[width=\imw]{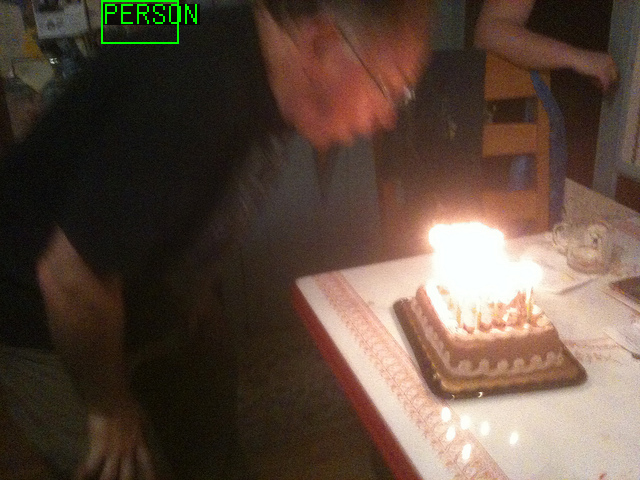}\\
    \vspace*{10pt}
    \includegraphics[width=\imw]{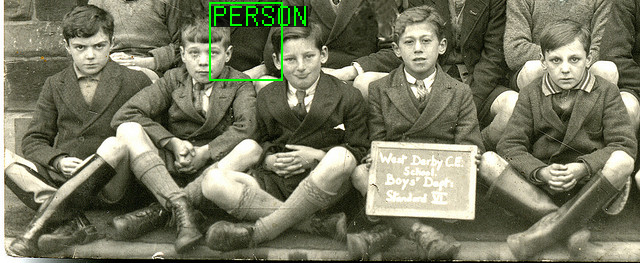}\\
    \vspace*{10pt}
    \includegraphics[width=\imw]{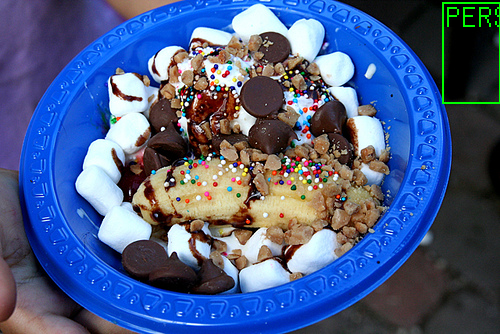}\\
    \vspace*{10pt}
    %\includegraphics[width=\imw]{img/coco-sup/4.7a-000001728150-spurious.png}\\
    %\vspace*{10pt}
    %\includegraphics[width=\imw]{img/coco-sup/5.7a-000002152486-spurious.png}\\
    %\vspace*{10pt}
    \includegraphics[width=\imw]{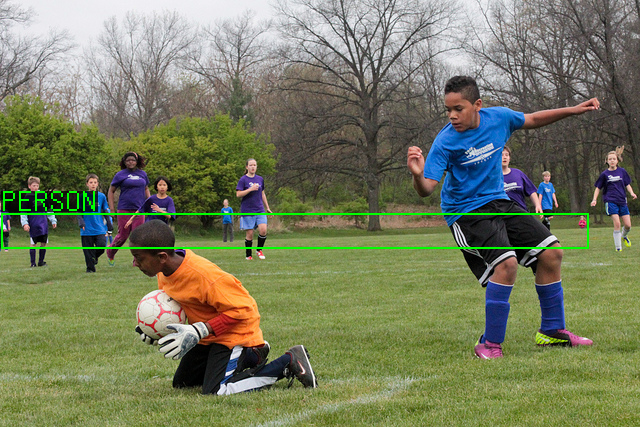}\\
    \vspace*{10pt}
    %\includegraphics[width=\imw]{img/coco-sup/10.3a-900100328430-spurious.png}\\
    %\vspace*{10pt}
    %\includegraphics[width=\imw]{img/coco-sup/16.3a-000000563639-spurious.png}\\
    %\vspace*{10pt}
    \includegraphics[width=\imw]{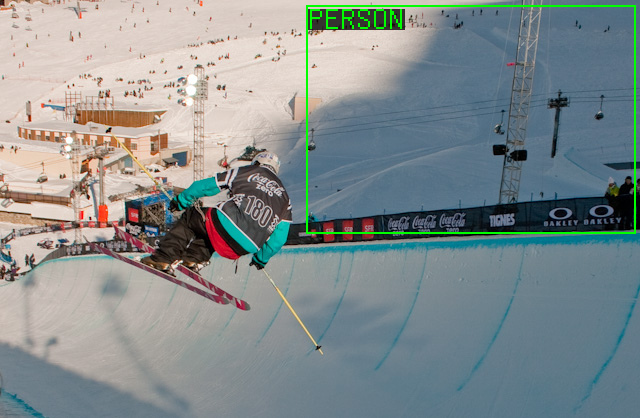}\\
    \caption{Spurious boxes, COCO.}
    \label{fig:examples-coco-spurious}
\end{figure}

Some instances of objects are blurry.
Examples are shown in Fig.~\ref{fig:examples-coco-blurry}.
Such objects probably should not have been labeled as they might be very challenging to detect.

\renewcommand\imw{0.69\linewidth}
\begin{figure}[phtb]
    \centering
    \includegraphics[width=\imw]{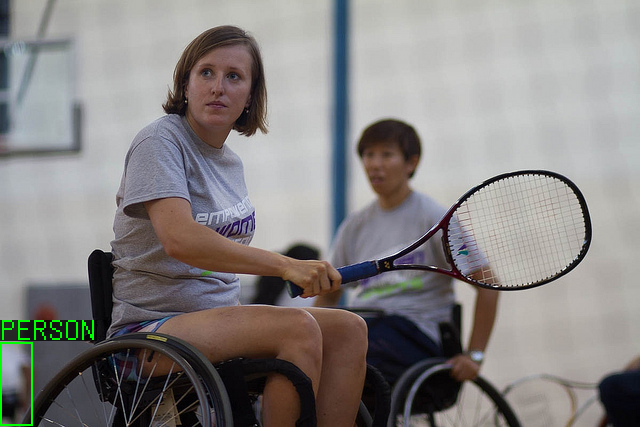}\\
    \vspace*{10pt}
    %\includegraphics[width=\imw]{img/coco-sup/1.1a-000001740597-blur.png}\\
    %\vspace*{10pt}
    \includegraphics[width=\imw]{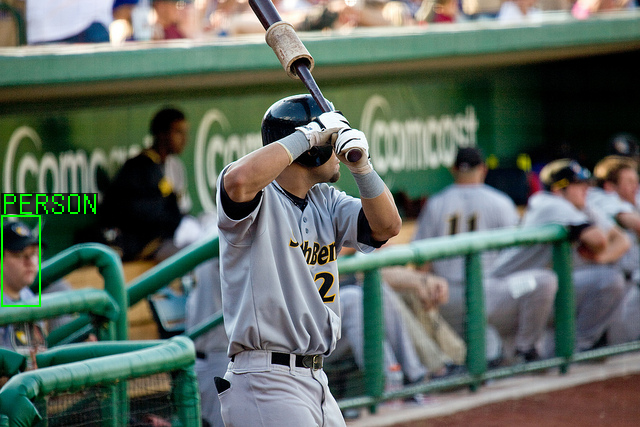}\\
    \vspace*{10pt}
    \includegraphics[width=\imw]{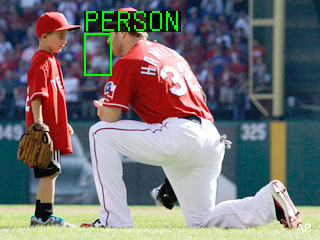}\\
    \vspace*{10pt}
    %\includegraphics[width=\imw]{img/coco-sup/2.1a-000001754246-blur.png}\\
    %\vspace*{10pt}
    %\includegraphics[width=\imw]{img/coco-sup/2.2a-000002207611-blur.png}\\
    %\vspace*{10pt}
    %\includegraphics[width=\imw]{img/coco-sup/4.0a-000001688680-blur.png}\\
    %\vspace*{10pt}
    \includegraphics[width=\imw]{img/annots/coco-2026707.png}\\
    \vspace*{10pt}
    \includegraphics[width=\imw]{img/annots/coco-223888.png}\\
    %\vspace*{10pt}
    %\includegraphics[width=\imw]{img/coco-sup/16.2a-000001709663-blur.png}\\
    %\vspace*{10pt}
    \caption{Blurry objects, COCO.}
    \label{fig:examples-coco-blurry}
\end{figure}

The last categories of boxes we found in the COCO dataset were incorrectly placed and mislabeled
ones (Fig.~\ref{fig:examples-coco-location}).
An interesting example, presented at the bottom of the figure, contains an ambiguous object that an annotator classified as \textit{truck}.

\renewcommand\imw{0.69\linewidth}
\begin{figure}[phtb]
    \centering
    \includegraphics[width=\imw]{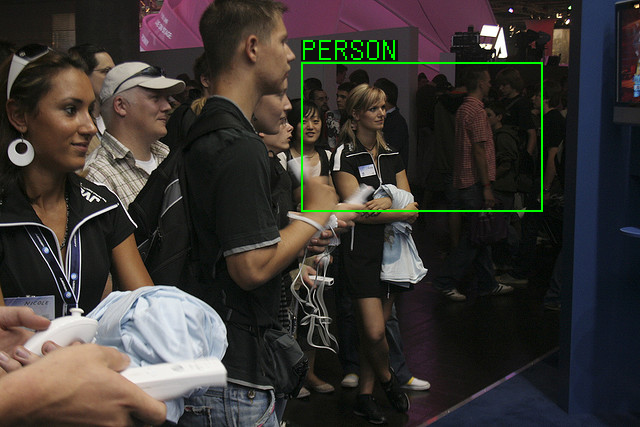}\\
    \vspace*{10pt}
    \includegraphics[width=\imw]{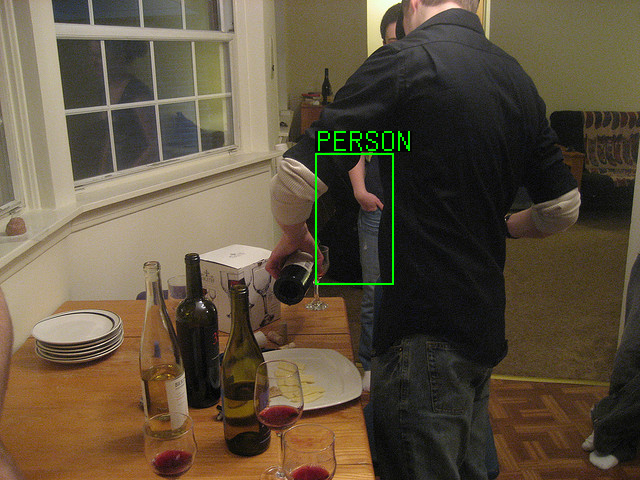}\\
    \vspace*{10pt}
    %\includegraphics[width=\imw]{img/annots/coco-2154394.png}\\
    %\vspace*{10pt}
    %\includegraphics[width=\imw]{img/annots/coco-1689689.png}\\
    %\vspace*{10pt}
    \includegraphics[width=\imw]{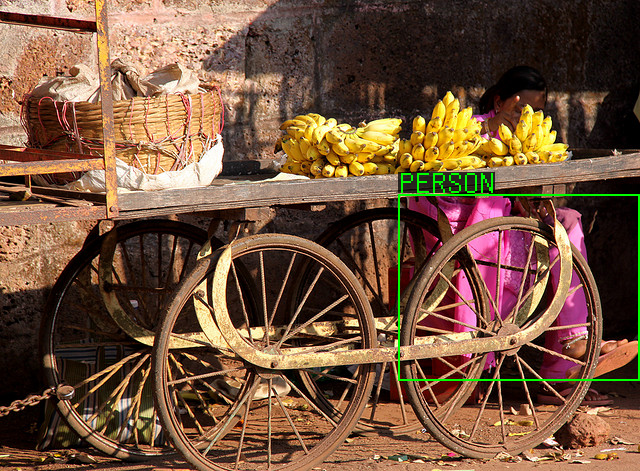}\\
    \vspace*{10pt}
    \includegraphics[width=\imw]{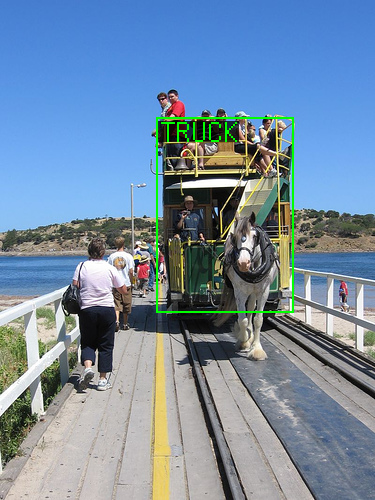}\\
    \caption{Examples of mislocated boxes (first three) and mislabeled boxes (bottom), COCO.}
    \label{fig:examples-coco-location}
\end{figure}

\iffalse
\renewcommand\imw{0.707\linewidth}
\begin{figure}[phtb]
    \centering
    \includegraphics[width=\imw]{img/coco-sup/10.4a-000000561905-parts.png}\\
    \vspace*{10pt}
    \includegraphics[width=\imw]{img/coco-sup/23.8a-000000517787-parts.png}\\
    \vspace*{10pt}
    \includegraphics[width=\imw]{img/coco-sup/30.1a-000000204277-parts.png}\\
    \vspace*{10pt}
    \includegraphics[width=\imw]{img/coco-sup/46.3a-000001203965-parts.png}\\
    \caption{Examples of labeled parts of objects, COCO.}
    \label{fig:examples-coco-parts}
\end{figure}
\fi

\begin{figure}[phtb]
    \centering
    \includegraphics[width=\imw]{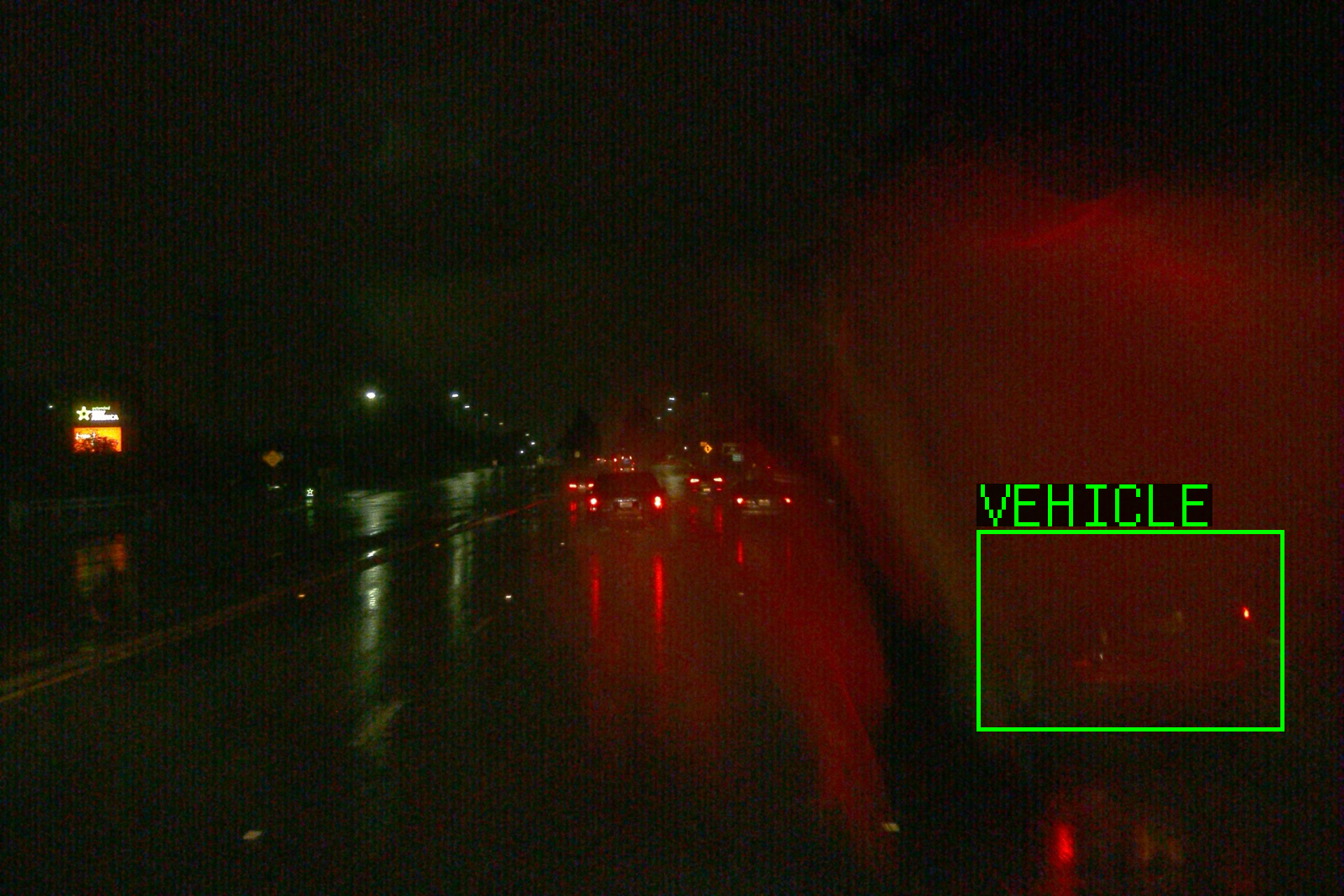}\\
    \vspace*{10pt}
    \includegraphics[width=\imw]{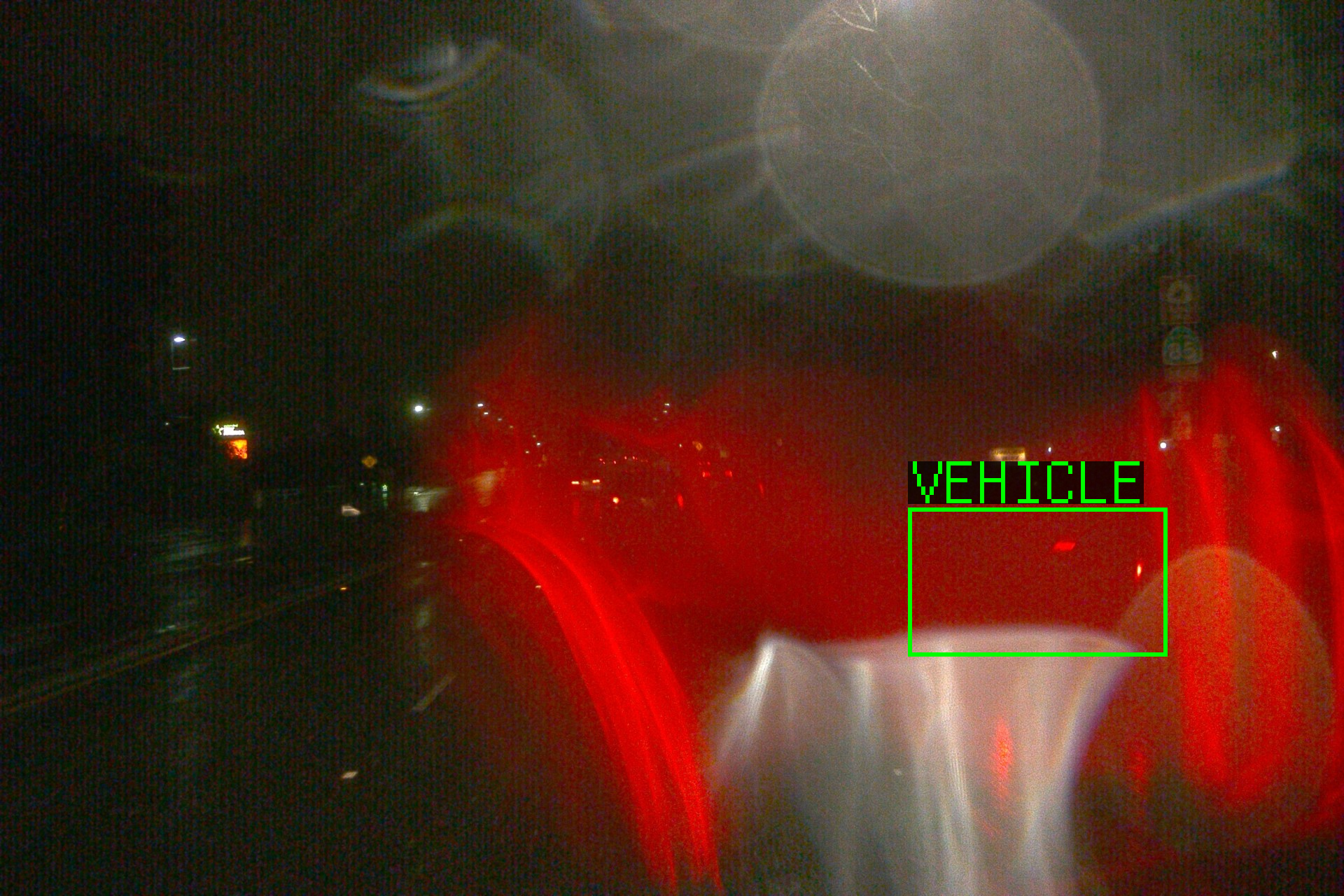}\\
    \caption{Objects hardly visible due to water on lens, Waymo.}
    \label{fig:examples-waymo-wet}
\end{figure}

\begin{figure}[phtb]
    \centering
    \includegraphics[width=\imw]{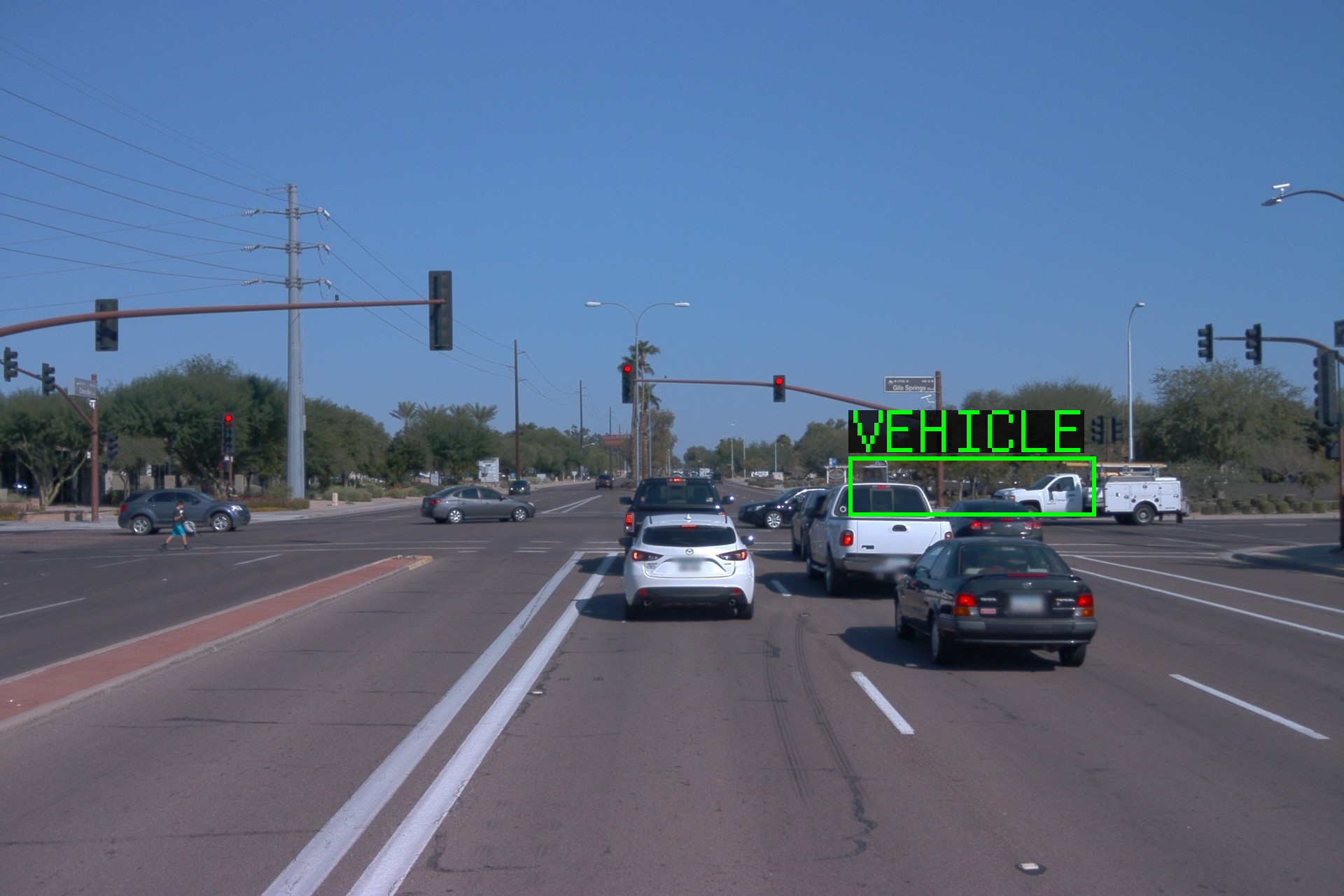}\\
    \vspace*{10pt}
    \includegraphics[width=\imw]{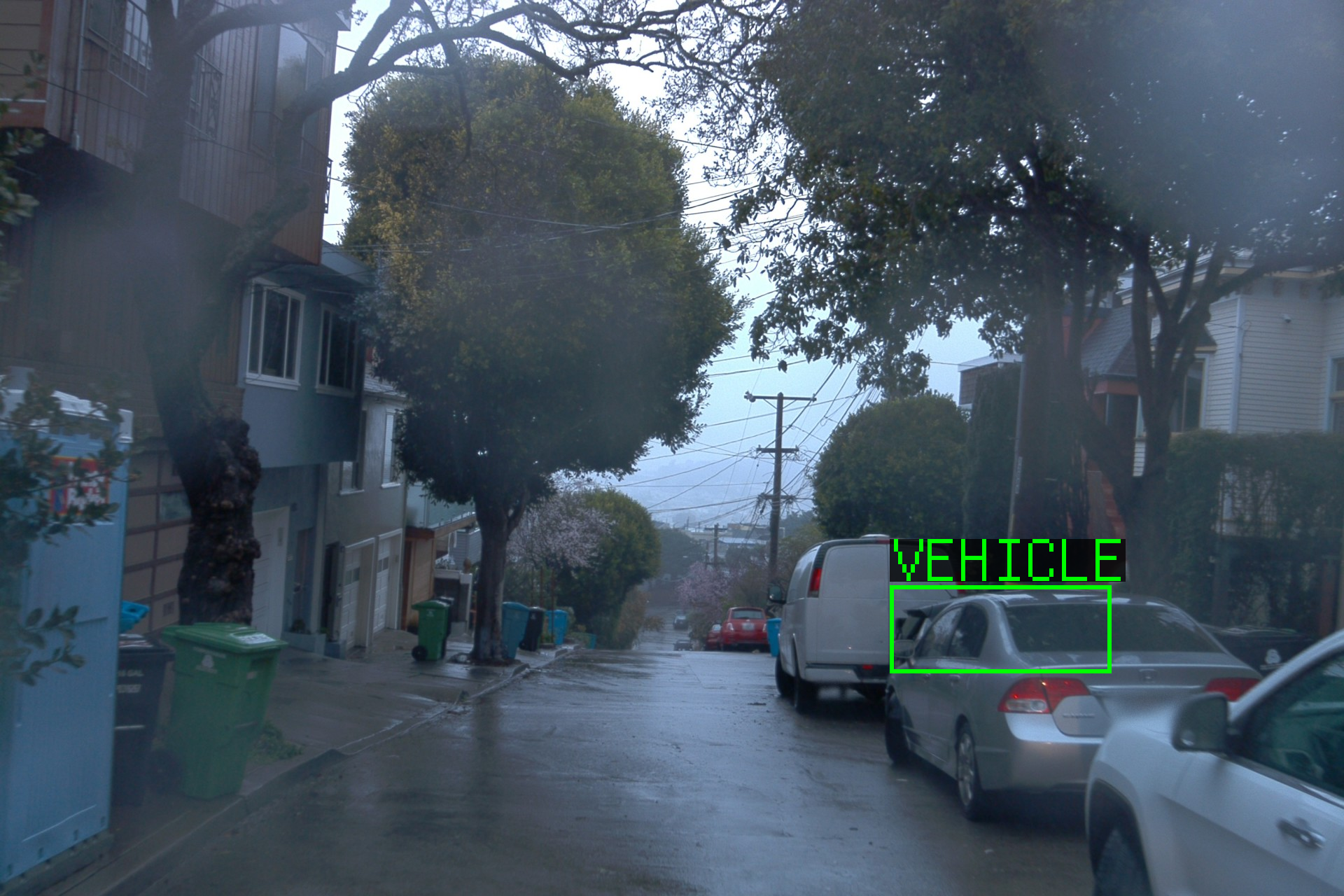}\\
    \caption{Incorrectly placed boxes, Waymo.}
    \label{fig:examples-waymo-mislocated}
\end{figure}

\begin{figure}[phtb]
    \centering
    \includegraphics[width=\imw]{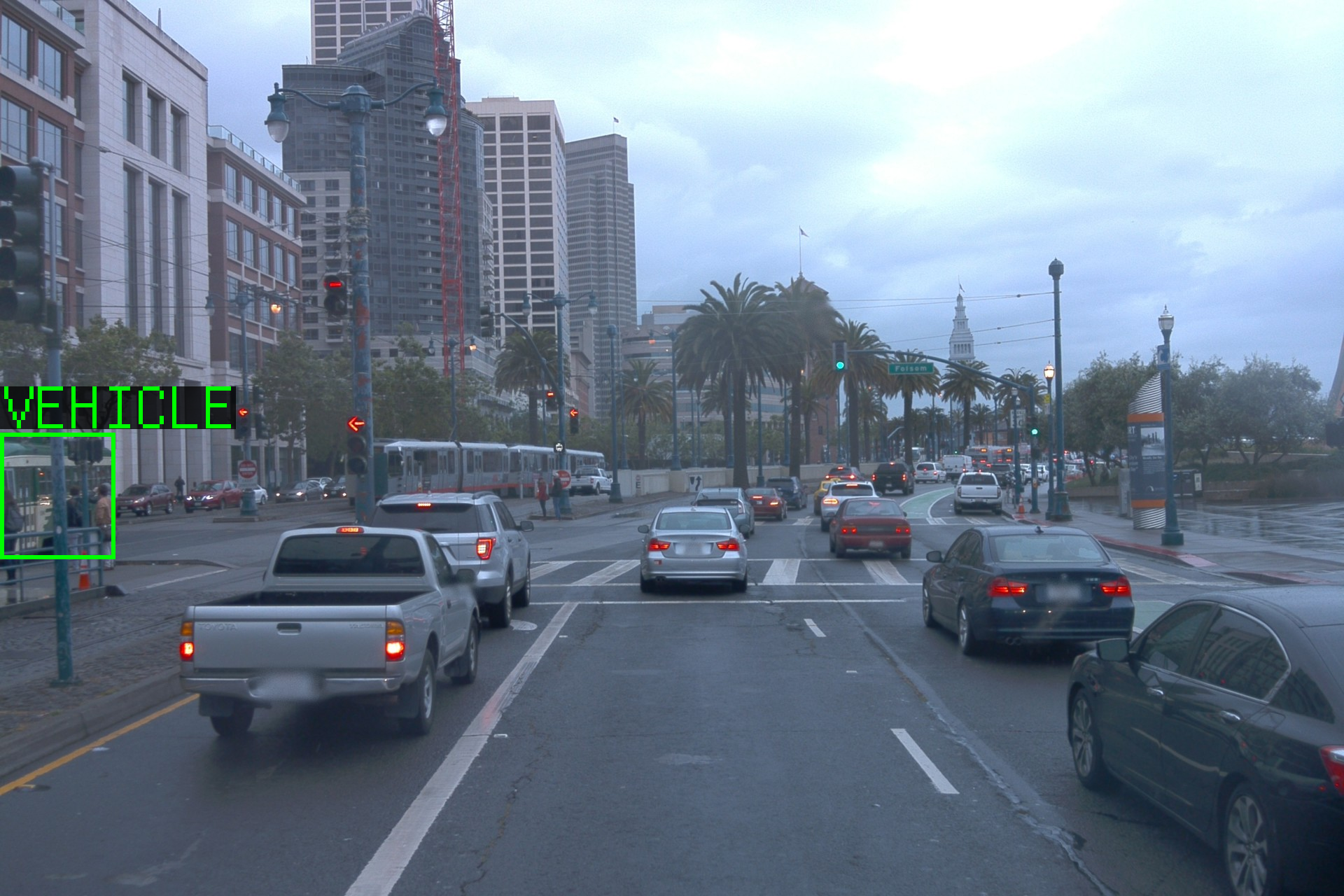}\\
    \vspace*{10pt}
    \includegraphics[width=\imw]{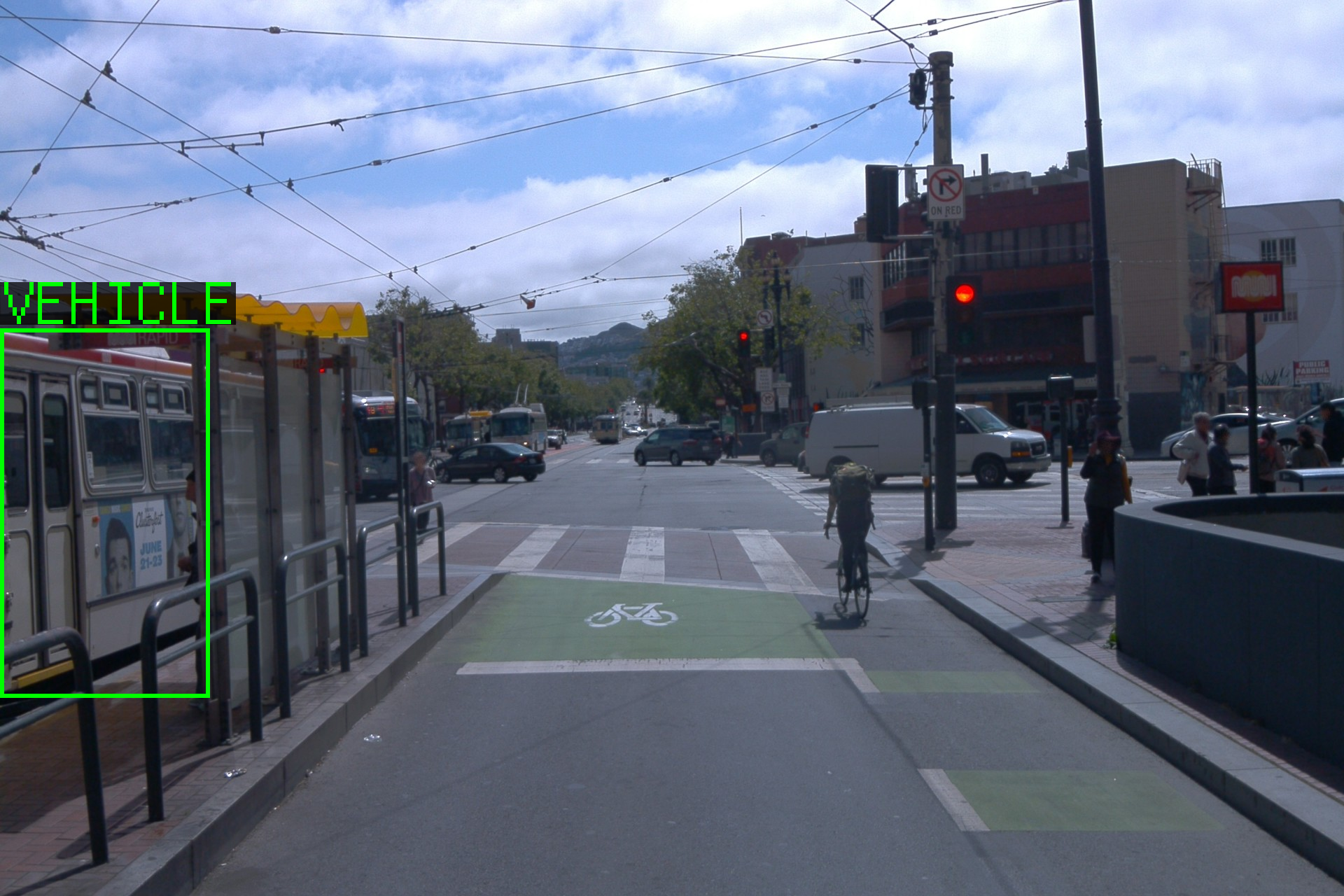}\\
    \caption{Objects that might not be vehicles, Waymo.}
    \label{fig:examples-waymo-notvehicles}
\end{figure}

In the Waymo dataset we found objects which are difficult to identify because of water on the camera lens (Fig.~\ref{fig:examples-waymo-wet}), some mislocated boxes (Fig.~\ref{fig:examples-waymo-mislocated}), boxes that do not appear to contain any objects (Fig.~\ref{fig:examples-coco-location}) and objects difficult to classify (Fig.~\ref{fig:examples-waymo-notvehicles}).

\begin{figure}[phtb]
    \centering
    \includegraphics[width=\imw]{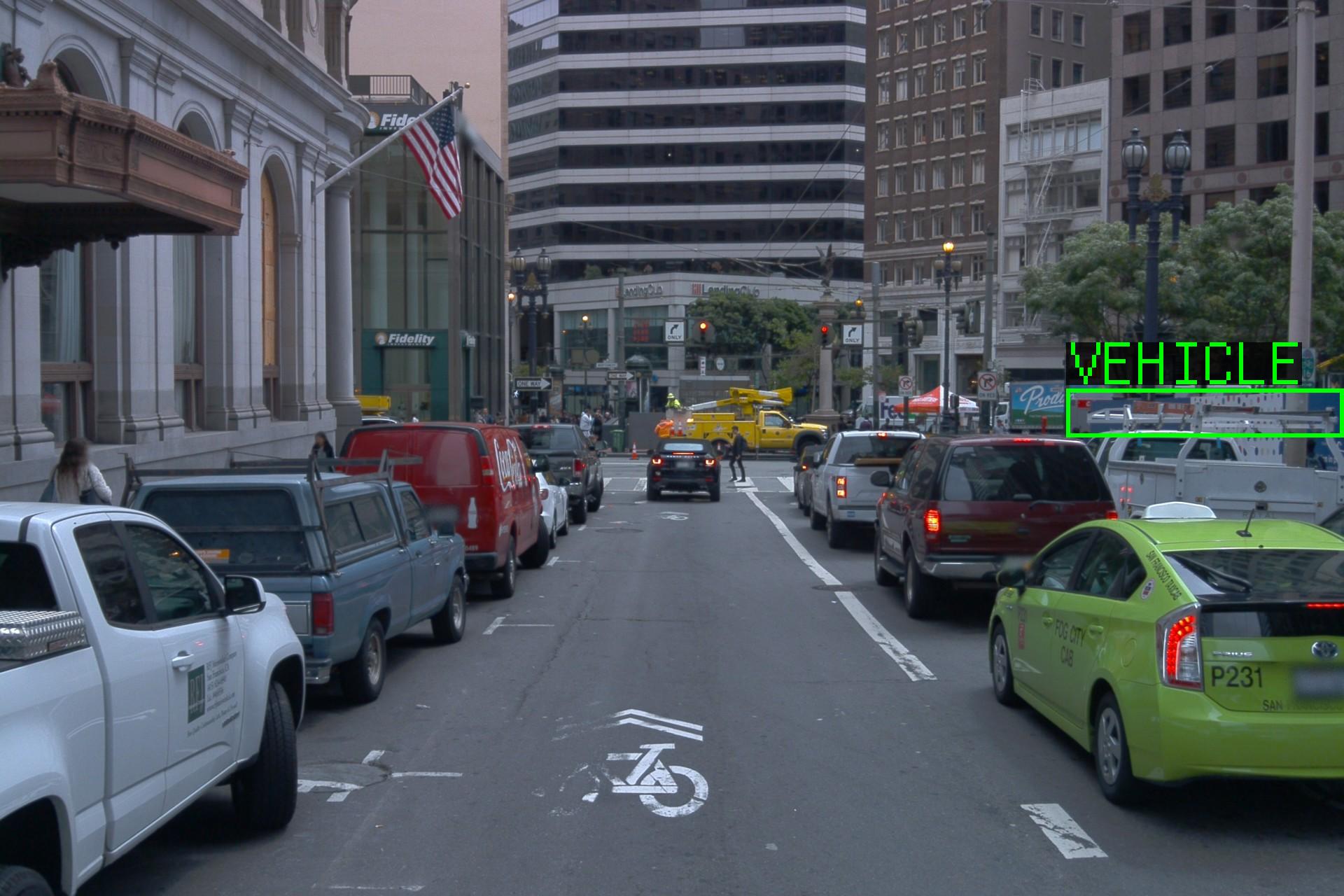}\\
    \vspace*{10pt}
    \includegraphics[width=\imw]{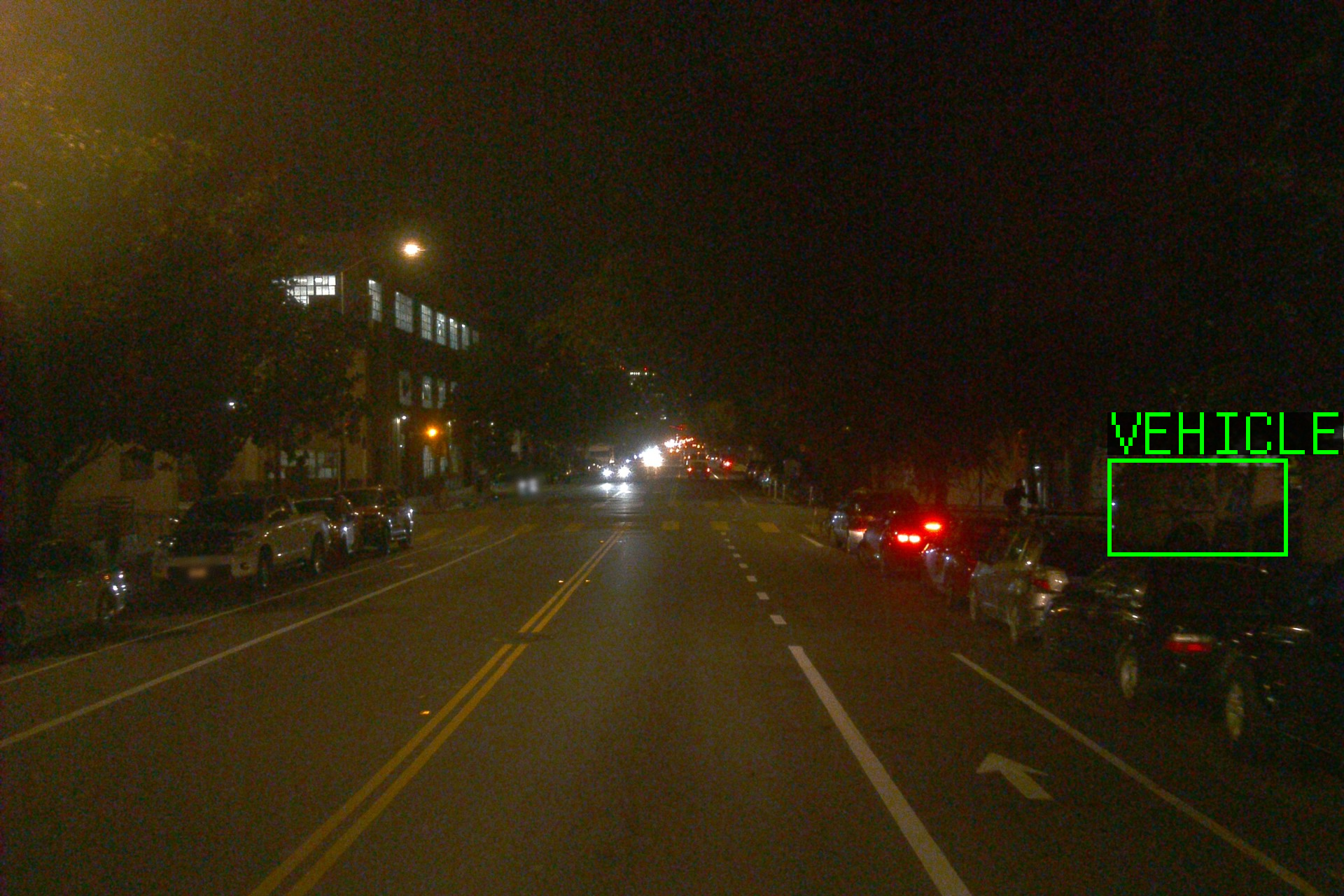}\\
    \caption{Ambiguous boxes, Waymo.}
    \label{fig:examples-waymo-location}
\end{figure}

%\renewcommand\imw{0.7\linewidth}
%\begin{figure}[phtb]
%    \centering
%    \includegraphics[width=\imw, keepaspectratio]{img/waymo-sup/9.4a-000000000170-nonconformant.png}
%    \caption{Protruding vertical part, Waymo.}
%    \label{fig:examples-waymo-crane}
%\end{figure}

% ------------------------------------------------------------------------------

\section{Discussion}

All of the experiments show that CLOD is a viable method for finding incorrect labels in object detection datasets.
It is the first method that finds and identifies missing, spurious, mislocated, and mislabeled
bounding boxes.
Thanks to the suggestion feature, in most cases, dataset reviewers only need to decide whether
to keep each suggestion or discard it.
Such an approach makes the whole process take less time than traditional
relabeling.

{\bf \noindent Noise impact.} The results of the first experiment have confirmed that a high-quality dataset
is crucial to obtain a high-quality model.
We parameterized the location and scale noise types, where the parameter specifies the
magnitude of the noise, i.e., how much the location and scale are changed.
During the first experiments, we set these values at 25\%, which
caused a relatively small impact on model quality.
Increasing this amplitude would have angled these curves further downwards.
Eliminating a number of noisy examples equal to 5\% of the dataset
could improve mAP@50 by as much as 0.085, which in some cases equates to 20\% improvement.
These results reveal the need for fixing erroneous labels in object detection datasets.

{\bf \noindent CLOD effectiveness.} The second experiment has shown that the method can detect all types of noise with
notable correctness as confirmed by the AUROC scores.
When it comes to disturbed location or scale of bounding boxes, low
displacements and scale factors were difficult to detect.
On the other hand, CLOD detected boxes with location and scale disturbed by 50\% of their
dimensions as often as those with other types of noise.

{\bf \noindent CLOD vs state-of-the-art.} Comparison of CLOD to ObjectLab yields favorable results for our new method.
It provides an additional benefit of detecting spurious boxes while also being
more effective in detecting missing ones.
However, the ObjectLab might be a better choice when focusing on detecting mislabeled
boxes in small, uniform datasets with few possible labels.
It's also worth mentioning that while CLOD completed the rating of a given dataset
in less time than ObjectLab (39.61 s vs. 42.29 s for COCO and 15.44 s vs. 37.68 s for Waymo), it used around twice as much system memory (921.55 MiB s vs 416.93 MiB for COCO and 432.77 MiB s vs 219.54 MiB for Waymo).

{\bf \noindent Dataset variants.} The similarity or decrease of mAP scores for revised datasets might be caused
by annotation practices that make the object detection task more difficult. Depending on the labeling policy, when very small objects are annotated, we may increase the complexity of the dataset for the model. Therefore, improving quality of the dataset, we can also make it more challenging for the model and decrease the original mAP metric, when the dataset was revised.
Jiaxin Ma et al. observed a similar effect in \cite{baobab}.
The high score of CLOD suggestions could be explained by model architecture
bias.
Both CLOD scoring and final training used the Faster R-CNN model.
More experiments must be performed to confirm this where, for these two
tasks, two different model architectures are used.
The \textit{selected} dataset variants always scored between revised and suggested variants.
That might indicate that some differences in the revised datasets are
not improvements but the opposite.

{\bf \noindent Manual inspection.} Suspicious examples found in original datasets show that the proposed CLOD method can be
utilized to find real labeling issues.
Browsing low-quality bounding boxes revealed that annotation guidelines are
inconsistent or not respected by annotators.
In the COCO dataset, there were many images where groups of small objects were
marked with a single large box.
In the Waymo dataset, bounding boxes often contained almost entirely obscured objects.
This is probably due to the fact that this dataset's original form is multi-camera video sequences
and point clouds.
It means that annotators saw more than single frames from the front camera.
Nonetheless, one might argue that a subset of an object detection dataset should still be
a valid dataset.

Unfortunately, the problem of noisy labels in object detection databases is
weakly addressed in the literature. 
The state-of-the-art datasets could contain a non-negligible amount of erroneous 
annotations. Therefore, the results obtained using such data can be considerably biased. 
There were initiatives to fix popular datasets in the past, like in \cite{baobab} and \cite{sama-coco}
but only parts of them were manually reviewed and re-annotated.
Another example could be \cite{MathiasBenenson2014} in which the authors proved that the evaluation
of existing face datasets is biased due to different guidelines for the annotators, and reviewing annotations could significantly improve the evaluation protocol of object detectors.
Datasets could be automatically searched using our method to find
suspicious annotations.
Then, only the images selected by CLOD could be manually re-annotated,
thus reducing the cost of such a project.

\section{Conclusion}

We have introduced the CLOD algorithm that successfully extends
the confident learning method to the domain of object detection.
In the experimental study, we have shown that it 
obtains promising results when applied to the task of evaluating annotations
in object detection datasets. 
Finding incorrect annotations with the achieved accuracy means that it is
viable to re-annotate all examples marked by the algorithm as suspicious.
The performed experiments proved that state-of-the-art object detection databases contain a non-negligible amount of noisy labels that decrease the maximal performance of the existing object detection architectures. It is probable that by removing noisy labels from these datasets, we could improve the quality and evaluation protocol of well-known
publicly available object detection datasets.
Moreover, the proposed methodology may improve the trustworthiness of the training datasets for
object detection algorithms used in safety-critical applications.

The CLOD source code is available under GNU General Public License v3 here: \url{https://github.com/safednn-group/safednn-clean}.

\section*{Acknowledgements}
This work was supported by the National Centre for Research and Development under the project LIDER/51/0221/L-11/19/NCBR/2020.
\balance
% ------------------------------------------------------------------------------

\todo[inline]{This note should be no further than at the top of page 13. This is page \thepage.}

\bibliographystyle{IEEEtran}
\bibliography{main}

\begin{thebibliography}{10}
\providecommand{\url}[1]{#1}
\csname url@samestyle\endcsname
\providecommand{\newblock}{\relax}
\providecommand{\bibinfo}[2]{#2}
\providecommand{\BIBentrySTDinterwordspacing}{\spaceskip=0pt\relax}
\providecommand{\BIBentryALTinterwordstretchfactor}{4}
\providecommand{\BIBentryALTinterwordspacing}{\spaceskip=\fontdimen2\font plus
\BIBentryALTinterwordstretchfactor\fontdimen3\font minus
  \fontdimen4\font\relax}
\providecommand{\BIBforeignlanguage}[2]{{%
\expandafter\ifx\csname l@#1\endcsname\relax
\typeout{** WARNING: IEEEtran.bst: No hyphenation pattern has been}%
\typeout{** loaded for the language `#1'. Using the pattern for}%
\typeout{** the default language instead.}%
\else
\language=\csname l@#1\endcsname
\fi
#2}}
\providecommand{\BIBdecl}{\relax}
\BIBdecl

\bibitem{Popowicz2022CombatingLN}
A.~Popowicz, K.~Radlak, S.~Lasota, K.~Szczepankiewicz, and M.~Szczepankiewicz,
  ``Combating label noise in image data using multinet flexible confident
  learning,'' \emph{Applied Sciences}, 2022.

\bibitem{ChangAmershi2017}
J.~C. Chang, S.~Amershi, and E.~Kamar, ``Revolt: Collaborative crowdsourcing
  for labeling machine learning datasets,'' in \emph{Proc. of the 2017 CHI
  Conference on Human Factors in Computing Systems}, 2017, p. 2334–2346.

\bibitem{WillersSudholt2020}
O.~Willers, S.~Sudholt, S.~Raafatnia, and S.~Abrecht, ``Safety concerns and
  mitigation approaches regarding the use of deep learning in safety-critical
  perception tasks,'' in \emph{Computer Safety, Reliability, and Security.
  SAFECOMP 2020 Workshops}, 2020, pp. 336--350.

\bibitem{Frnay2014ACI}
B.~Fr{\'e}nay and A.~Kab{\'a}n, ``A comprehensive introduction to label
  noise,'' in \emph{The European Symposium on Artificial Neural Networks},
  2014.

\bibitem{ISO-21448}
\BIBentryALTinterwordspacing
I.~O. for Standardization, ``{ISO/PAS} 21448 -- {R}oad {V}ehicles - {S}afety of
  the intended functionality,'' 2022. [Online]. Available:
  \url{https://www.iso.org/standard/77490.html}
\BIBentrySTDinterwordspacing

\bibitem{ISO-4804}
\BIBentryALTinterwordspacing
------, ``{ISO/TR} 4804 {R}oad vehicles — {S}afety and cybersecurity for
  automated driving systems — design, verification and validation,'' 2020.
  [Online]. Available: \url{https://www.iso.org/standard/80363.html}
\BIBentrySTDinterwordspacing

\bibitem{UL4600}
\BIBentryALTinterwordspacing
{UL Standards \& Engagement}, ``{UL 4600}, standard for evaluation of
  autonomous products, {E}dition~1.'' [Online]. Available:
  \url{https://ul.org/UL4600}
\BIBentrySTDinterwordspacing

\bibitem{Northcutt2021PervasiveLE}
C.~G. Northcutt, A.~Athalye, and J.~Mueller, ``Pervasive label errors in test
  sets destabilize machine learning benchmarks,'' \emph{ArXiv}, vol.
  abs/2103.14749, 2021.

\bibitem{Northcutt2021ConfidentLE}
C.~G. Northcutt, L.~Jiang, and I.~L. Chuang, ``Confident learning: Estimating
  uncertainty in dataset labels,'' \emph{J. Artif. Intell. Res.}, vol.~70, pp.
  1373--1411, 2021.

\bibitem{Thyagarajan2023MultilabelCL}
A.~Thyagarajan, E.~Snorrason, C.~Northcutt, and J.~Mueller, ``Identifying
  incorrect annotations in multi-label classification data,'' in \emph{ICLR
  Workshop on Trustworthy ML}, 2023.

\bibitem{LampertStumpf2016}
T.~A. Lampert, A.~Stumpf, and P.~Gançarski, ``An empirical study into
  annotator agreement, ground truth estimation, and algorithm evaluation,''
  \emph{IEEE Transactions on Image Processing}, vol.~25, no.~6, pp. 2557--2572,
  2016.

\bibitem{Reed2015TrainingDN}
S.~E. Reed, H.~Lee, D.~Anguelov, C.~Szegedy, D.~Erhan, and A.~Rabinovich,
  ``Training deep neural networks on noisy labels with bootstrapping,''
  \emph{CoRR}, vol. abs/1412.6596, 2015.

\bibitem{Goldberger2017TrainingDN}
J.~Goldberger and E.~Ben-Reuven, ``Training deep neural-networks using a noise
  adaptation layer,'' in \emph{ICLR}, 2017.

\bibitem{Jiang2018MentorNetLD}
L.~Jiang, Z.~Zhou, T.~Leung, L.-J. Li, and L.~Fei-Fei, ``Mentornet: Learning
  data-driven curriculum for very deep neural networks on corrupted labels,''
  in \emph{ICML}, 2018.

\bibitem{Chen2019UnderstandingAU}
P.~Chen, B.~Liao, G.~Chen, and S.~Zhang, ``Understanding and utilizing deep
  neural networks trained with noisy labels,'' in \emph{ICML}, 2019.

\bibitem{Han2018CoteachingRT}
B.~Han, Q.~Yao, X.~Yu, G.~Niu, M.~Xu, W.~Hu, I.~Tsang, and M.~Sugiyama,
  ``Co-teaching: Robust training of deep neural networks with extremely noisy
  labels,'' in \emph{NeurIPS}, 2018.

\bibitem{Chadwick2019TrainingOD}
S.~Chadwick and P.~Newman, ``Training object detectors with noisy data,''
  \emph{2019 IEEE Intelligent Vehicles Symposium (IV)}, pp. 1319--1325, 2019.

\bibitem{Mao2021NoisyAR}
J.~Mao, Q.~Yu, Y.~Yamakata, and K.~Aizawa, ``Noisy annotation refinement for
  object detection,'' in \emph{BMVC}, 2021.

\bibitem{Liu2022TowardsRA}
X.~Liu, W.~Li, Q.~Yang, B.~Li, and Y.~Yuan, ``Towards robust adaptive object
  detection under noisy annotations,'' \emph{IEEE/CVF Conf. on Computer Vision
  and Pattern Recognition (CVPR)}, pp. 14\,187--14\,196, 2022.

\bibitem{Wang2022NoisyLocationAnnotations}
\BIBentryALTinterwordspacing
S.~Wang, J.~Gao, B.~Li, and W.~Hu, ``Narrowing the gap: Improved detector
  training with noisy location annotations,'' \emph{{IEEE} Transactions on
  Image Processing}, vol.~31, pp. 6369--6380, 2022. [Online]. Available:
  \url{https://doi.org/10.1109\%2Ftip.2022.3211468}
\BIBentrySTDinterwordspacing

\bibitem{objectlab}
U.~Tkachenko, A.~Thyagarajan, and J.~Mueller, ``Objectlab: Automated diagnosis
  of mislabeled images in object detection data,'' in \emph{ICML Workshop on
  Data-centric Machine Learning}, 2023.

\bibitem{HAC}
F.~Nielsen, \emph{Hierarchical Clustering}.\hskip 1em plus 0.5em minus
  0.4em\relax Springer International Publishing, 02 2016, pp. 195--211.

\bibitem{sama-coco}
{Samasource Impact Sourcing}, ``The sama-coco dataset,''
  \url{https://www.sama.com/sama-coco-dataset/}, 2022, accessed: 2023-04-28.

\bibitem{Ren2015FasterRT}
S.~Ren, K.~He, R.~Girshick, and J.~Sun, ``Faster {R-CNN}: Towards real-time
  object detection with region proposal networks,'' in \emph{Proc. of the 28th
  International Conference on Neural Information Processing Systems - Volume
  1}, ser. NIPS'15, 2015, p. 91–99.

\bibitem{Lin2017FocalLf}
T.-Y. Lin, P.~Goyal, R.~Girshick, K.~He, and P.~Dollár, ``Focal loss for dense
  object detection,'' in \emph{2017 IEEE Int. Conf. on Computer Vision (ICCV)},
  2017, pp. 2999--3007.

\bibitem{Redmon2018YAI}
\BIBentryALTinterwordspacing
J.~Redmon and A.~Farhadi, ``Yolov3: An incremental improvement,'' 2018.
  [Online]. Available: \url{https://arxiv.org/abs/1804.02767}
\BIBentrySTDinterwordspacing

\bibitem{Li2020TowardsNO}
\BIBentryALTinterwordspacing
J.~Li, C.~Xiong, R.~Socher, and S.~Hoi, ``Towards noise-resistant object
  detection with noisy annotations,'' 2020. [Online]. Available:
  \url{https://arxiv.org/abs/2003.01285}
\BIBentrySTDinterwordspacing

\bibitem{baobab}
J.~Ma, Y.~Ushiku, and M.~Sagara, ``The effect of improving annotation quality
  on object detection datasets: A preliminary study,'' in \emph{Proceedings of
  the IEEE/CVF Conference on Computer Vision and Pattern Recognition}, 2022,
  pp. 4850--4859.

\bibitem{MathiasBenenson2014}
M.~Mathias, R.~Benenson, M.~Pedersoli, and L.~Van~Gool, ``Face detection
  without bells and whistles,'' in \emph{Computer Vision -- ECCV 2014}, 2014,
  pp. 720--735.

\end{thebibliography}

\end{document}